\documentclass[10pt,twocolumn,letterpaper]{article}

\usepackage{cvpr}
\usepackage{times}
\usepackage{epsfig}
\usepackage{graphicx}
\usepackage{amsmath}
\usepackage{amssymb}
\usepackage{booktabs}
\usepackage{paralist}
\usepackage{multirow}
\usepackage{enumitem}
\usepackage{subcaption}
\usepackage{wrapfig}
\usepackage{mysymbols}

\newcommand{\myquote}[1]{\emph{`#1'}}
\newcommand{\myapprox}{{\raise.17ex\hbox{$\scriptstyle\sim$}}}
\newcommand{\xhdr}[1]{\vspace{2pt}\noindent\textbf{#1}}

\usepackage[toc,page]{appendix}
\usepackage[usenames,dvipsnames]{xcolor}
\usepackage{changepage}   
\usepackage{xcolor}
\usepackage{colortbl}

\usepackage{amssymb}
\usepackage{pifont}
\newcommand{\cmark}{\ding{51}}%
\newcommand{\xmark}{\ding{55}}%

\newcommand{\squintguy}[0]{\includegraphics[width=.04\textwidth]{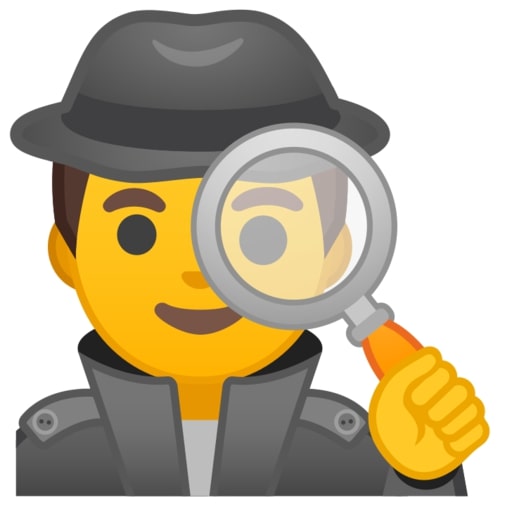}}

\definecolor{orange}{rgb}{1,0.5,0}
\definecolor{lightsalmonpink}{rgb}{1.0, 0.6, 0.6}
\definecolor{verylightsalmonpink}{rgb}{0.966, 0.805, 0.797}
\definecolor{lightblue}{rgb}{0.862, 0.906, 0.984}
\definecolor{lightyellow}{rgb}{1.0, 0.945, 0.797}
\definecolor{lightgreen}{rgb}{0.835, 0.91, 0.828}
\definecolor{lightpurple}{rgb}{0.879, 0.832, 0.902}

\usepackage{array}
\newcolumntype{P}[1]{>{\centering\arraybackslash}p{#1}}

\usepackage[pagebackref=true,breaklinks=true,letterpaper=true,colorlinks,bookmarks=false]{hyperref}
\hypersetup{
    colorlinks = true
    }

\cvprfinalcopy 

\def\cvprPaperID{1939} 

\newcommand{\data}{VQA-introspect}
\newcommand{\perc}{Perception}
\newcommand{\reas}{Reasoning}
\newcommand{\squint}{SQuINT}

\ifcvprfinal\fi

\belowdisplayskip \abovedisplayskip

\newlength{\abstractReduceTop}
\newlength{\abstractReduceBot}

\newcommand{\reducedSection}[1]{\vspace{\sectionReduceTop}\section{#1}\vspace{\sectionReduceBot}}
\newlength{\sectionReduceTop}
\newlength{\sectionReduceBot}

\newlength{\subsectionReduceTop}
\newlength{\subsectionReduceBot}

\newlength{\subsubsectionReduceTop}
\newlength{\subsubsectionReduceBot}

\newlength{\captionReduceTop}
\newlength{\captionReduceBot}

\newlength{\eqnReduceTop}
\newlength{\eqnReduceBot}

\newlength{\horSkip}
\newlength{\verSkip}

\newlength{\figureHeight}
\begin{document}

\title{\squintguy~ SQuINTing at VQA Models: \\Introspecting VQA Models with Sub-Questions}

\author{
	Ramprasaath R. Selvaraju$^1$\thanks{Research performed in part during an internship at Microsoft Research} \hspace{1pc}
	Purva Tendulkar$^1$ \hspace{1pc}
    Devi Parikh$^{1}$ \hspace{1pc}\\
    Eric Horvitz$^{2}$ \hspace{1pc}
    Marco Tulio Ribeiro$^{2}$ \hspace{1pc}
    Besmira Nushi$^{2}$ \hspace{1pc}
    Ece Kamar$^{2}$ \hspace{1pc}\\
    \hspace{-1pc}$^1$Georgia Institute of Technology,
	$^2$Microsoft Research\\
	{\tt\small \{ramprs, purva, parikh\}@gatech.edu}\\
	{\tt\small \{horvitz, marcotcr, benushi, eckamar\}@microsoft.com}\\
}

\maketitle

\thispagestyle{plain}
\pagestyle{plain}

\begin{abstract}

\vspace{-8pt} 
Existing VQA datasets contain questions with varying levels of complexity. While the majority of questions in these datasets require \emph{perception} for recognizing existence, properties, and spatial relationships of entities, a significant portion of questions pose challenges that correspond to \emph{reasoning tasks} -- tasks that can only be answered through a synthesis of perception and knowledge about the world, logic and / or reasoning. 
Analyzing performance across this distinction allows us to notice when existing VQA models have consistency issues; they answer the reasoning questions correctly but fail on associated low-level perception questions. 
For example, in Figure \ref{fig:squint_teaser}, models answer the complex reasoning question ``Is the banana ripe enough to eat?" correctly, but fail on the associated perception question ``Are the bananas mostly green or yellow?'' indicating that the model likely answered the reasoning question correctly but for the wrong reason. 
We quantify the extent to which this phenomenon occurs by creating a new \reas~split of the VQA dataset and collecting \data{}, a new dataset\footnote{Our dataset can be found at \href{http://aka.ms/vqa-introspect}{aka.ms/vqa-introspect}.} which consists of 238K new perception questions which serve as sub questions corresponding to the set of perceptual tasks needed to effectively answer the complex reasoning questions in the \reas~split. 
Our evaluation shows that state-of-the-art VQA models have comparable performance in answering perception and reasoning questions, but suffer from consistency problems. 
To address this shortcoming, we propose an approach called Sub-Question Importance-aware Network Tuning (SQuINT), which encourages the model to attend to the same parts of the image when answering the reasoning question and the perception sub question.
We show that SQuINT improves model consistency by $\sim$5\%, also marginally improving performance on the \reas~questions in VQA, while also displaying better attention maps. 

\end{abstract}

\reducedSection{Introduction}

\begin{figure}[t]
  \centering
  \includegraphics[width=0.95\columnwidth]{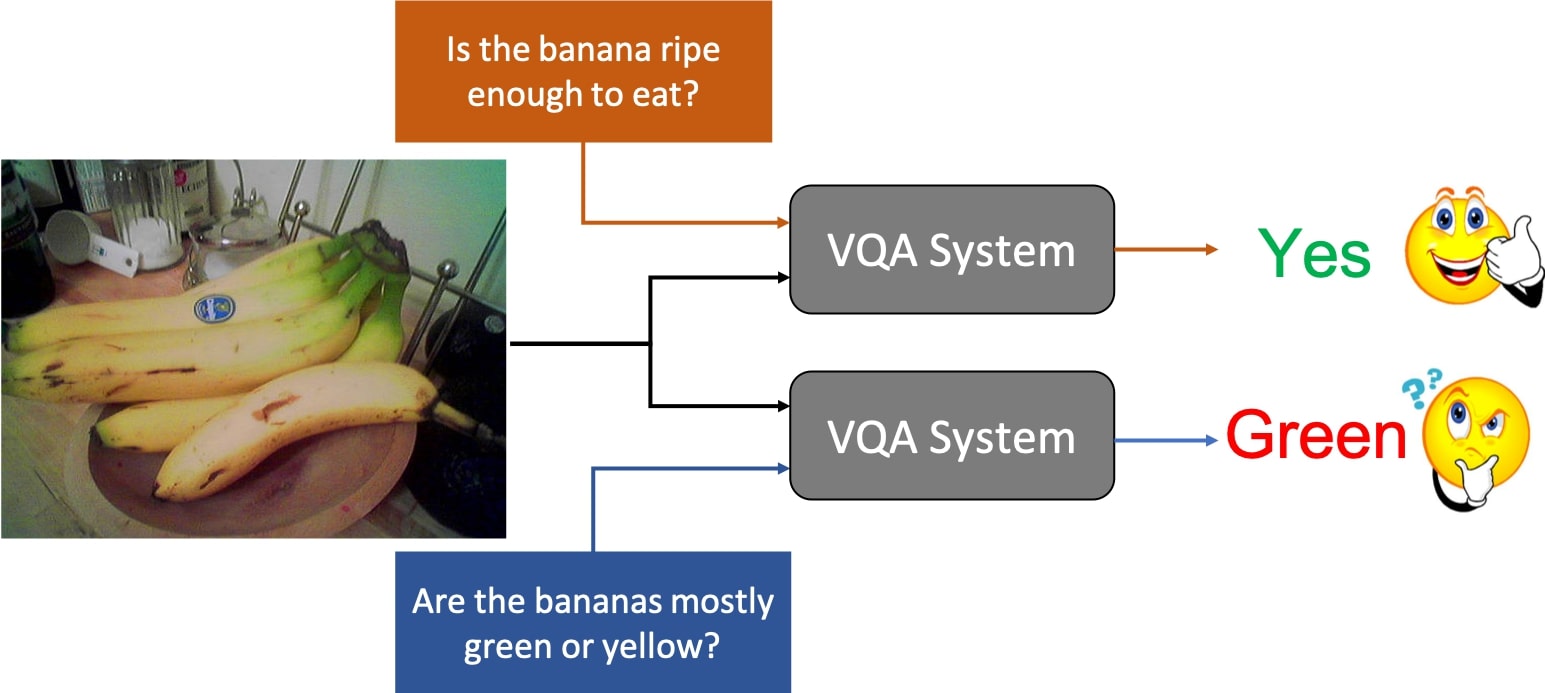}
   \caption{A potential reasoning failure: Current models answer the \reas~question ``Is the banana ripe enough to eat?'' correctly with the answer ``Yes''. We might assume that doing so stems from perceiving relevant concepts correctly -- perceiving yellow bananas in this example. But when asked ``Are the bananas mostly green or yellow?'', the model answers the question incorrectly with ``Green'' -- indicating that the model possibly answered the original Reasoning question for the wrong reasons even if the answer was right. We quantify the extent to which this phenomenon occurs in VQA and introduce a new dataset aimed at stimulating research on well-grounded reasoning.}
  \label{fig:squint_teaser}
\vspace{-10pt}
\end{figure}

Human cognition is thought to be compositional in nature: the visual system recognizes multiple aspects of a scene which are combined into shapes \cite{vision_is_compositional} and understandings. Likewise, complex linguistic expressions are built from simpler ones \cite{language_is_compositional}.  
Similarly, tasks like Visual Question Answering (VQA) require models to perform inference at multiple levels of abstraction.
For example, to answer the question, ``Is the banana ripe enough to eat?'' (Figure~\ref{fig:squint_teaser}), a VQA model has to be able to detect the bananas and extract associated properties such as size and color (perception), understand what the question is asking, and reason about how these properties relate to known properties of edible bananas (ripeness) and how they manifest (yellow versus green in color).
While ``abstraction'' is complex and spans distinctions at multiple levels of detail, we focus on separating questions into  \perc~and \reas~questions. \perc~questions only require visual perception to recognize existence, physical properties or spatial relationships among entities, such as ``What color is the banana?'' or ``What is to the left of the man?'', while \reas~questions require the composition of multiple perceptual tasks and knowledge that harnesses logic and prior knowledge about the world, such as ``Is the banana ripe enough to eat?''.

Current VQA datasets \cite{antol2015vqa, goyal2016making, ren2015exploring} contain a mixture of \perc~and \reas~questions, which are considered equivalent for the purposes of evaluation and learning.
Categorizing questions into \perc~and \reas~ promises to promote a better assessment of visual perception and higher-level reasoning capabilities of models, rather than conflating these capabilities.
Furthermore, we believe it is useful to identify the \perc~questions that serve as subtasks in the compositional processes required to answer the \reas~question. By elucidating such ``sub-questions,'' we can check whether the model is reasoning appropriately or if it is relying on spurious shortcuts and biases in datasets \cite{agrawal2018don}. 
For example, we should be cautious about the model's inferential ability if it simultaneously answers ``no'' to ``Are the bananas edible?'' and ``yellow'' to ``What color are the bananas?'', even if the answer to the former question is correct. The inconsistency between the higher-level reasoning task and the lower-level perception task that it builds upon suggests that the system has not learned effectively how to answer the Reasoning question and will not be able to generalize to same or closely related \reas~question with another image. 
The fact that these sub-questions are in the same modality (i.e. questions with associated answers) allows for the evaluation of any VQA model, rather than only models that are trained to provide justifications. It is this key observation that we use to develop an evaluation methodology for Reasoning questions.

The dominant learning paradigm for teaching models to answer VQA tasks assumes that models are given $<$image, question, answer$>$ triplets, with no additional annotation on the relationship between the question and the compositional steps required to arrive at the answer. 
As reasoning questions become more complex, achieving good coverage and generalization with methods used to date will likely require a prohibitive amount of data. Alternatively, we employ a hierarchical decomposition strategy, where we  identify and link Reasoning questions with sets of appropriate Perception sub-questions.  Such an approach promises to enable new efficiencies via compositional modeling, as well as lead to improvements in the consistency of models for answering Reasoning questions. 
Explicitly representing dependencies between \reas~tasks and the corresponding \perc~tasks also provides language-based grounding for reasoning questions where visual grounding \cite{qiao2018exploring, Selvaraju_2019_ICCV} may be insufficient, e.g., highlighting that the banana is important for the question in Figure~\ref{fig:squint_teaser} does not tell the model how it is important (i.e. that color is an important property rather than size or shape).
Again, the fact that such grounding is in question-answer form (which models already have to deal with) is an added benefit.
Such annotations allow for attempts to enforce reasoning devoid of shortcuts that do not generalize, or are not in line with human values and business rules, even if accurate (e.g. racist behavior).

We propose a new split of the VQA dataset, containing only \reas~questions (defined previously). Furthermore, for questions in the split, we introduce \data{}, a new dataset of 238K associated \perc~ sub-questions which humans perceive as containing the sub-questions needed to answer the original questions.
After validating the quality of the new dataset, we use it to perform fine-grained evaluation of state-of-the-art models, checking whether their reasoning is in line with their perception. We show that state-of-the-art VQA models have similar accuracy in answering perception and reasoning tasks but have problems with consistency; in 28.14\% of the cases where models answer the reasoning question correctly, they fail to answer the corresponding perception sub-question, highlighting problems with consistency and the risk that models may be learning to answer reasoning questions through learning common answers and biases. 

Finally, we introduce \squint~-- a generic modeling approach that is inspired by the compositional learning paradigm observed in humans. 
SQuINT incorporates  \data{} annotations into learning with a new loss function that encourages image regions important for the sub-questions to play a role in answering the main \reas~questions.  
Empirical evaluations demonstrate that the approach results in models that are more consistent across \reas~and associated \perc~tasks with no major loss of accuracy. 
We also find that SQuINT improves model attention maps for \reas~questions, thus making models more trustworthy.

\reducedSection{Related Work}
\label{sec:rel}

Visual Question Answering \cite{antol2015vqa}, one of the most widely studied vision-and-language problems, 
requires associating image content with natural language questions and answers (thus combining perception, language understanding, background knowledge and reasoning). 
However, it is possible for models to do well on the task by exploiting language and dataset biases, e.g. answering ``yellow'' to ``What color is the banana?'' without regard for the image or by answering ``yes'' to most yes-no questions \cite{agrawal2018don, li-tell, Selvaraju_2019_ICCV, YinYang, anne2018women}.
This motivates additional forms of evaluation, e.g. checking if the model can understand question rephrasings \cite{cycle_consistency} or whether it exhibits logical consistency \cite{redroses, ray2019sunny}. In this work, we present a novel evaluation of questions that require reasoning capabilities, where we check for consistency between how models answer higher level Reasoning questions and how they answer corresponding Perception sub-questions.

A variety of datasets have been released with attention annotations on the image pointing to regions that are important to answer questions (\cite{vqahat, huk2018multimodal}), with corresponding work on enforcing such grounding \cite{selvaraju2017grad, qiao2018exploring, Selvaraju_2019_ICCV}. Our work is complementary to these approaches, as we provide language-based grounding (rather than visual), and further evaluate the link between perception capabilities and how they are composed by models for answering Reasoning tasks.
Closer to our work is the dataset of Lisa et al. \cite{huk2018multimodal}, where natural language justifications are associated with (question, answer) pairs. However, most of the questions contemplated (like much of the VQA dataset) pertain to perception questions (e.g. for the question-answer ``What is the person doing? Snowboarding'', the justification is ``...they are on a snowboard ...''). Furthermore, it is hard to use natural language justifications to evaluate models that do not generate similar rationales (i.e. most SOTA models), or even coming up with metrics for models that do. In contrast, our dataset and evaluation is in the same modality (QA) that models are already trained to handle.

\section{Reasoning-VQA and \data{}} \label{dataset}


In the first part of this section, we present an analysis of the common type of questions in the VQA dataset and highlight the need for classifying them into \perc~and \reas~questions. 
We then define \perc~and \reas~questions and describe our method for constructing the \reas~split. 
In the second part, we describe how we create the new \data{} dataset through collecting sub-questions and answers for questions in our \reas~split. 
Finally, we describe experiments conducted in order to validate the quality of our collected data.

\subsection{\perc~vs. \reas}

A common technique for finer-grained evaluation of VQA models is to group instances by answer type (yes/no, number, other) or by the first words of the question (what color, how many, etc) \cite{antol2015vqa}.
While useful, such slices are coarse and do not evaluate the model's capabilities at different points in the abstraction scale. For example, questions like ``Is this a banana?'' and ``Is this a healthy food?'' start with the same words and expect yes/no answers.
While both test if the model can do object recognition, the latter requires additional capabilities in connecting recognition with prior knowledge about which food items are healthy and which are not.
This is not to say that \reas~questions are inherently harder, but that they require both visual understanding and an additional set of skills (logic, prior knowledge, etc) while \perc~questions deal mostly with visual understanding. For example, the question ``How many round yellow objects are to the right of the smallest square object in the image?'' requires very complicated visual understanding, and is arguably harder than ``Is the banana ripe enough to eat?'', which requires relatively simple visual understanding (color of the bananas) and knowledge about properties of ripe bananas. 
Regardless of difficulty, categorizing questions as Perception or Reasoning is useful for both detailed model evaluation based on capabilities and also improving learning, as we demonstrate in later sections. We now proceed to define these categories more formally.


\noindent\textbf{\perc~:} We define \perc~questions as those which can be answered by detecting and recognizing the existence, physical properties and / or spatial relationships between entities, recognizing text / symbols, simple activities and / or counting, and that do not require more than one hop of reasoning or general commonsense knowledge beyond what is visually present in the image. 
Some examples are: ``Is that a cat? '' (existence), ``Is the ball shiny?'' (physical property), ``What is next to the table?'' (spatial relationship), ``What does the sign say?'' (text / symbol recognition), ``Are the people looking at the camera?'' (simple activity), etc. We note that spatial relationship questions have been considered reasoning tasks in previous work~\cite{hudson2019gqa} as they require lower-level perception tasks in composition to be answered. For our purposes it is useful to separate visual understanding from other types of reasoning and knowledge, and thus we classify such spatial relationships as \perc.

\noindent\textbf{\reas~:} We define \reas~questions as non-\perc~questions which require the synthesis of perception with prior knowledge and / or reasoning in order to be answered.
For instance, ``Is this room finished or being built?'', ``At what time of the day would this meal be served?'', ``Does this water look fresh enough to drink?'', ``Is this a home or a hotel?'', ``Are the giraffes in their natural habitat?'' are all \reas~ questions. 

Our analysis of the perception questions in the VQA dataset revealed that most perception questions have distinct patterns that can be identified with high precision regex-based rules. By handcrafting such rules (details can be found in the Supplementary) and filtering out perception questions, we identify ~18\% of the VQA dataset as highly likely to be \reas. 
To check the accuracy of our rules and validate their coverage of \reas~questions, we designed a crowdsourcing task on Mechanical Turk that instructed workers to identify a given VQA question as \perc~or \reas, and to subsequently provide sub-questions for the \reas~questions, as described next. 
89.25\% of the times, at least 2 out of 3 workers classified our resulting questions as reasoning questions demonstrating the high precision of the regex-based rules we created. 

\subsection{\data{} data} 
Given the complexity of distinguishing between Perception / Reasoning and providing sub-questions for Reasoning questions, we first train and filter workers on Amazon Mechanical Turk (AMT) via qualification rounds before we rely on them to generate high-quality sub-questions.

\noindent\textbf{Worker Training - } \label{worker-training} We manually annotate $100$ questions from the VQA dataset as \perc{} and $100$ as \reas{} questions, to serve as examples.
We first teach crowdworkers the difference between \perc~and \reas~questions by presenting definitions and showing several examples of each, along with explanations. 
Then, crowdworkers are shown (question, answer) pairs and are asked to identify if the given question is a \perc~question or a \reas~question \footnote{We also add an ``Invalid'' category to flag nonsensical questions or those which can be answered without looking at the image}. 
Finally, for \reas~ questions, we ask workers to add all \perc~questions and corresponding answers (in short) that would be necessary to answer the main question (details and interface can be found in the Supplementary). 
In this qualification HIT, workers have to make 6 \perc{} and \reas{} judgments, and they qualify if they get 5 or more answers right.

We launched further pilot experiments for the crowdworkers who passed the first qualification round, where we manually evaluated the quality of their sub-questions based on whether they were \perc~questions grounded in the image and sufficient to answer the main question. 
Among those 540 workers who passed the first qualification test, 144 were selected (via manual evaluation) as high-quality workers, who finally qualified for attempting our main task.  

\noindent\textbf{Main task - } 
In the main data collection, all VQA questions identified as \reas~by regex-rules and a random subset of the questions identified as \perc~were further judged by workers (for validation purposes).  
We eliminated ambiguous questions by further filtering out questions where there is high worker disagreement about the answer. We required at least 8 out of 10 workers to agree with the majority answer for yes/no questions and 5 out of 10 for all other questions. This labeling step left us with a Reasoning split that corresponds to $\sim$13\% of the VQA dataset.

At the next step. each $<$question, image$>$ pair labeled as \reas~had sub questions generated by 3 unique workers \footnote{A small number of workers displayed degraded performance after the qualification round, and were manually filtered}. Removing duplicate question, answer pairs left on average $3.1$ sub-questions per \reas~question. 
Qualitative examples from the resulting dataset  are presented in Fig. \ref{fig:subq_data}.

The resulting train split of \data{} contains 166927 sub questions for 55799 \reas{} questions in VQAv2 train, and the val split of \data{} contains 71714 sub questions for 21677 \reas{} questions in VQAv2 val. 
This \reas~split is not exhaustive, but is high precision (as demonstrated below) and contains questions that are not ambiguous, and thus is useful for evaluation and learning.

\begin{figure*}[t!]
 \centering
 \begin{subfigure}{.5\textwidth}
  \centering
  \includegraphics[scale=0.12]{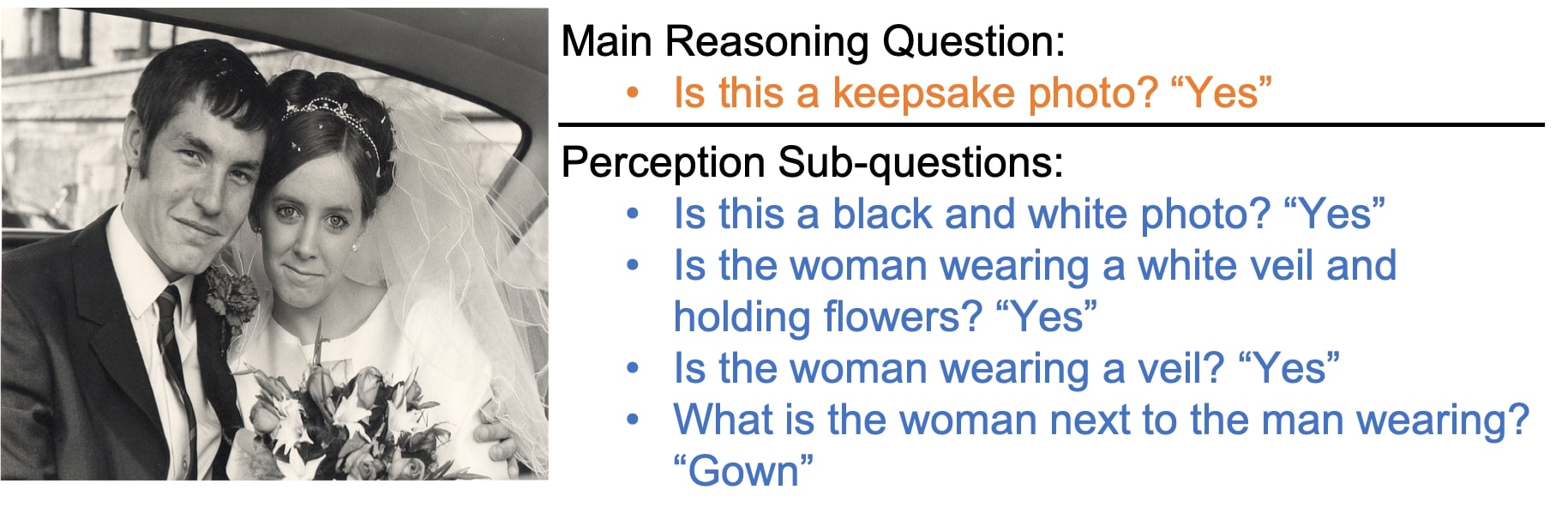}
  \caption{}
\end{subfigure}%
\begin{subfigure}{.5\textwidth}
  \centering
  \includegraphics[scale=0.12]{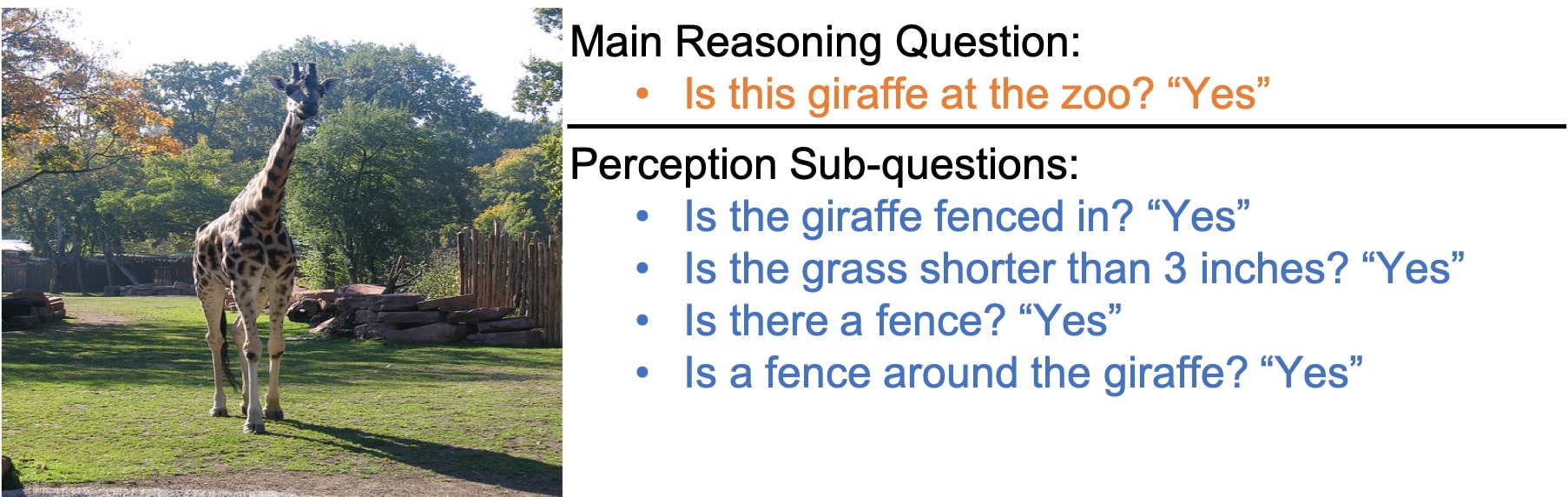}
  \caption{}
\end{subfigure}
\begin{subfigure}{.5\textwidth}
  \centering
  \includegraphics[scale=0.12]{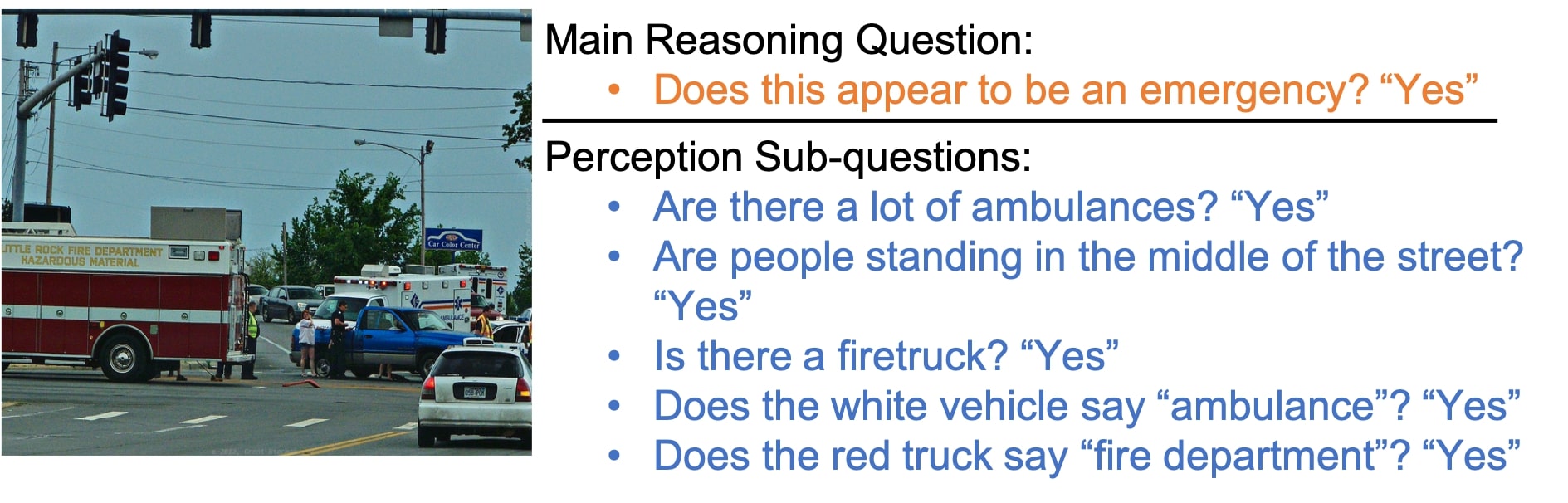}
  \caption{}
\end{subfigure}%
\begin{subfigure}{.5\textwidth}
  \centering
  \includegraphics[scale=0.12]{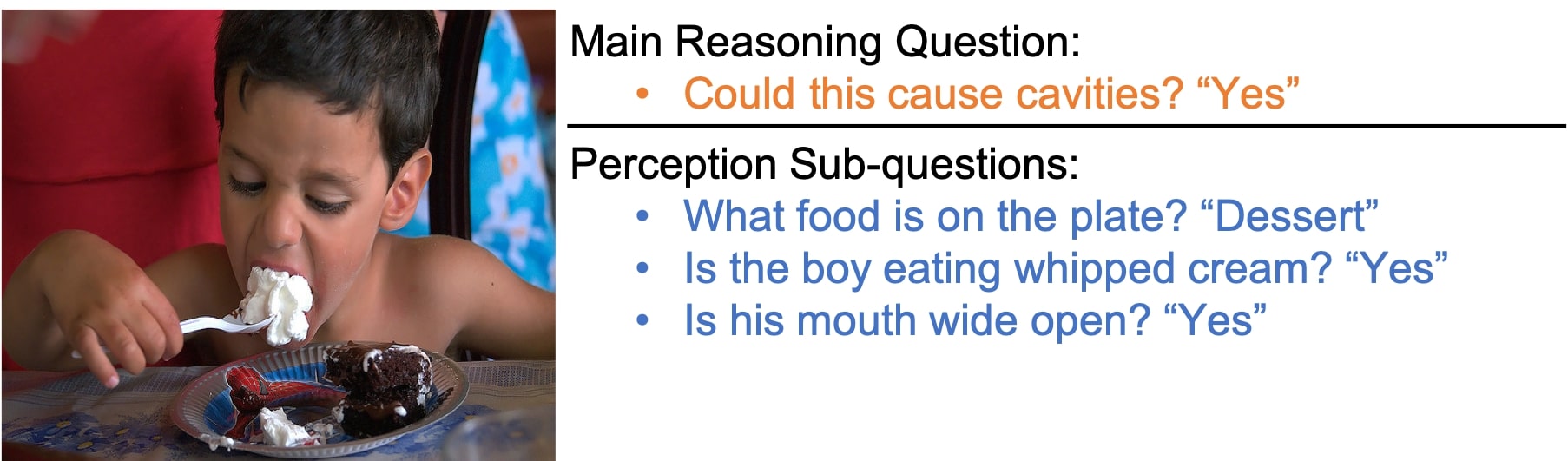}
  \caption{}
\end{subfigure}
\begin{subfigure}{.5\textwidth}
  \centering
  \includegraphics[scale=0.12]{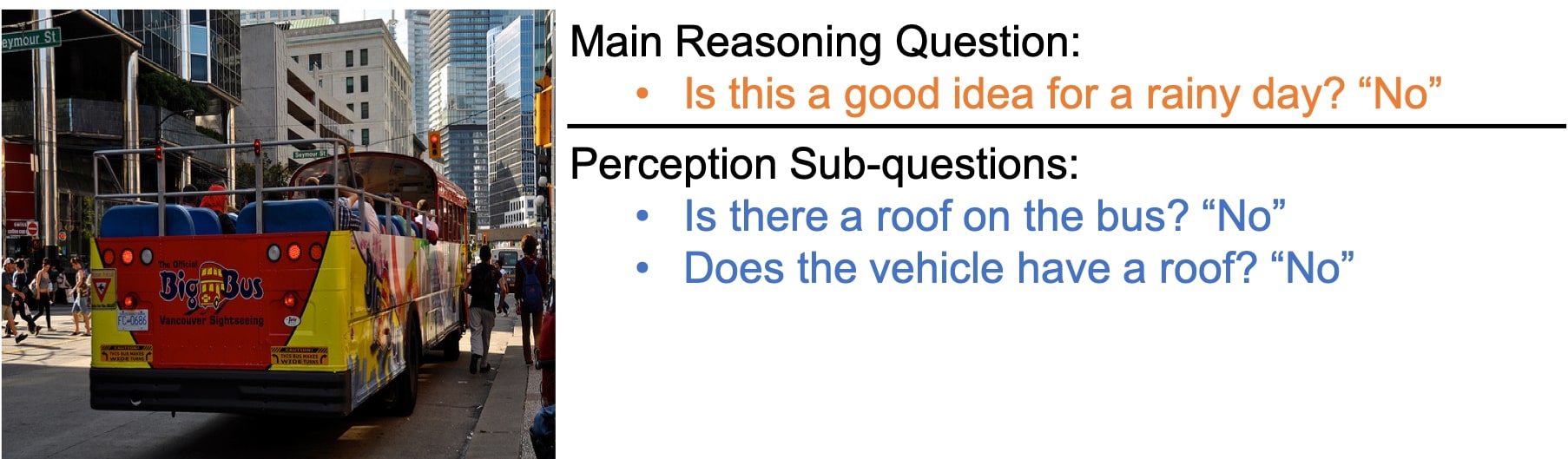}
  \caption{}
\end{subfigure}%
\begin{subfigure}{.5\textwidth}
  \centering
  \includegraphics[scale=0.12]{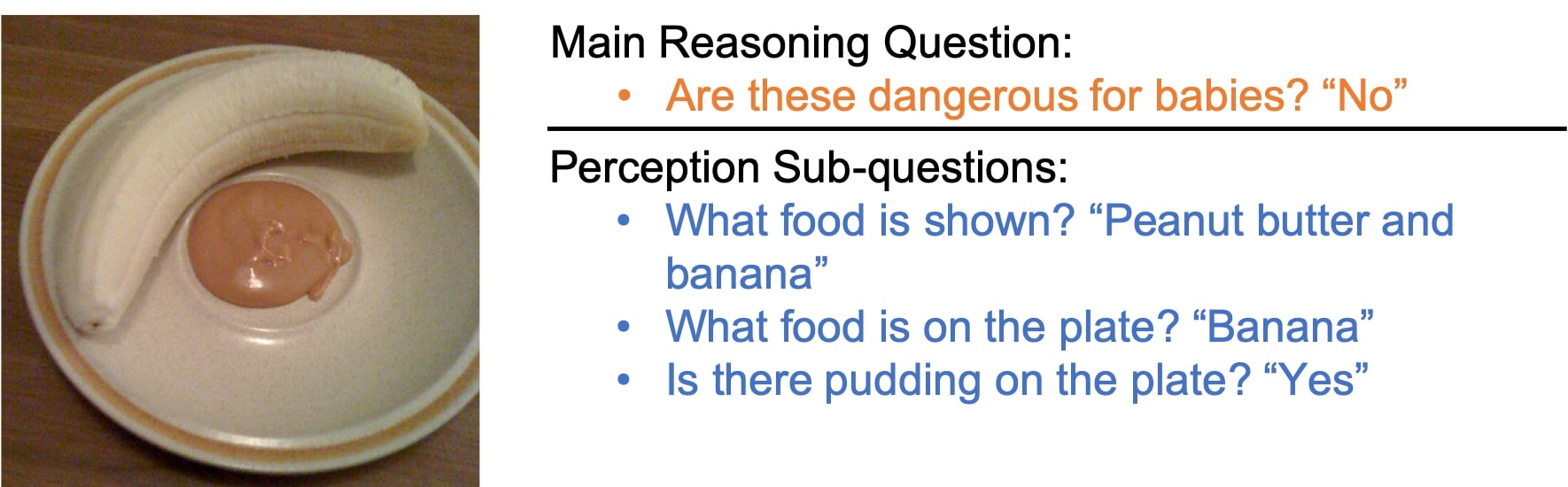}
  \caption{}
\end{subfigure}
\begin{subfigure}{.5\textwidth}
  \centering
  \includegraphics[scale=0.12]{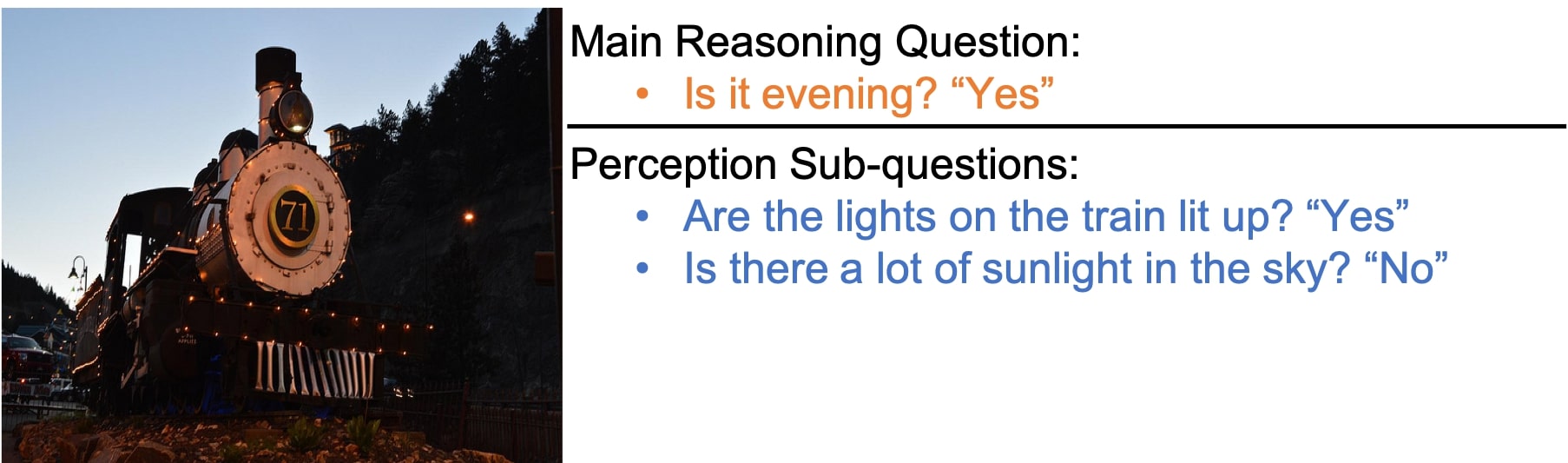}
  \caption{}
\end{subfigure}%
\begin{subfigure}{.5\textwidth}
  \centering
  \includegraphics[scale=0.12]{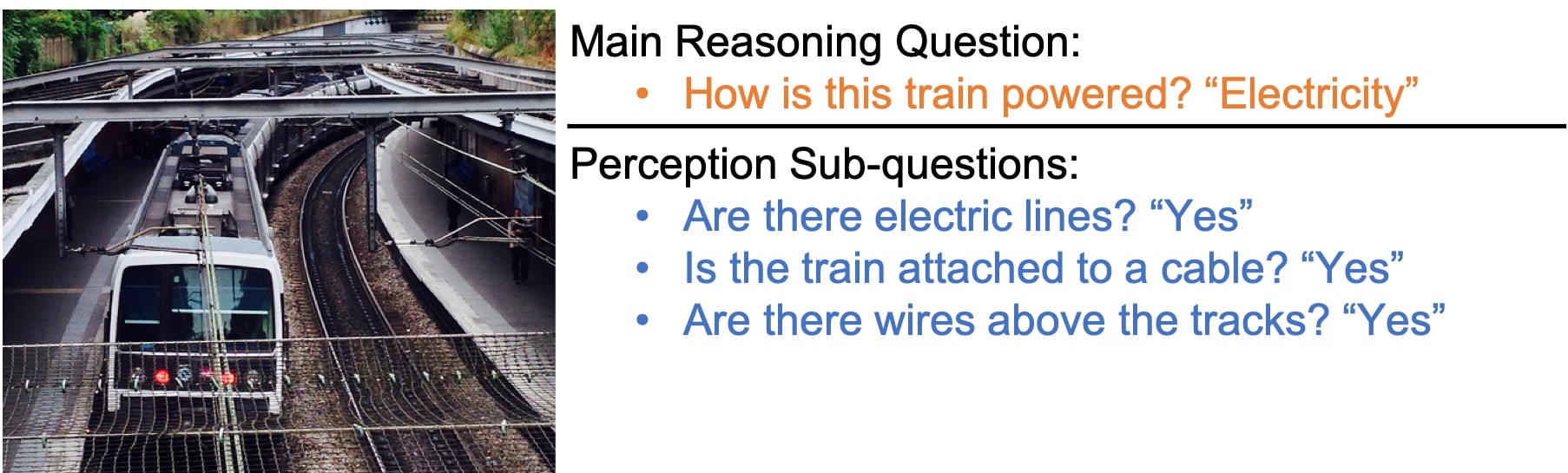}
  \caption{}
\end{subfigure}
\vspace{-10pt}
\caption{Qualitative examples of \perc~sub-questions in our \data{} dataset for main questions in the \reas~split of VQA. Main questions are in \textcolor{orange}{orange} and sub questions are in \textcolor{blue}{blue}. A single worker may have provided more than one sub questions for the same (image, main question) pair. More examples can be found in the Supplementary}
\vspace{-20pt}
\label{fig:subq_data}
\end{figure*}

\subsection{Dataset Quality Validation}

In order to confirm that the sub-questions in \data{} are really \perc{} questions, we did a further round of evaluation with workers who passed the worker qualification task described in Section~\ref{worker-training} but had not provided sub-questions for our main task. In this round, $87.8\%$ of sub-questions in \data{} were judged to be \perc{} questions by at least 2 out of 3 workers.

It is crucial for the semantics of \data{} that the sub-questions are tied to the original \reas{} question. While verifying that the sub-questions are necessary to answer the original question requires workers to think of all possible ways the original question could be answered (and is thus too hard), we devised an experiment to check if the sub-questions provide at least sufficient visual understanding to answer the \reas{} question. In this experiment, workers are shown the sub-questions with answers, and then asked to answer the \reas{} question without seeing the image, thus having to rely only on the visual knowledge conveyed by the sub-questions.
At least 2 out of 3 workers were able to answer 89.3\% of the \reas{} questions correctly in this regime (95.4\% of binary \reas{} questions). For comparison, when we asked workers to answer \reas{} questions with no visual knowledge at all (no image and no sub-questions), this accuracy was 52\% (58\% for binary questions).
These experiments give us confidence that the sub-questions in \data{} are indeed \perc{} questions that convey components of visual knowledge which can be composed to answer the original \reas{} questions.

\section{Dataset Analysis}

\begin{figure*}[t!]
\centering
 \begin{subfigure}{.35\textwidth}
   \centering
   \includegraphics[scale=0.12]{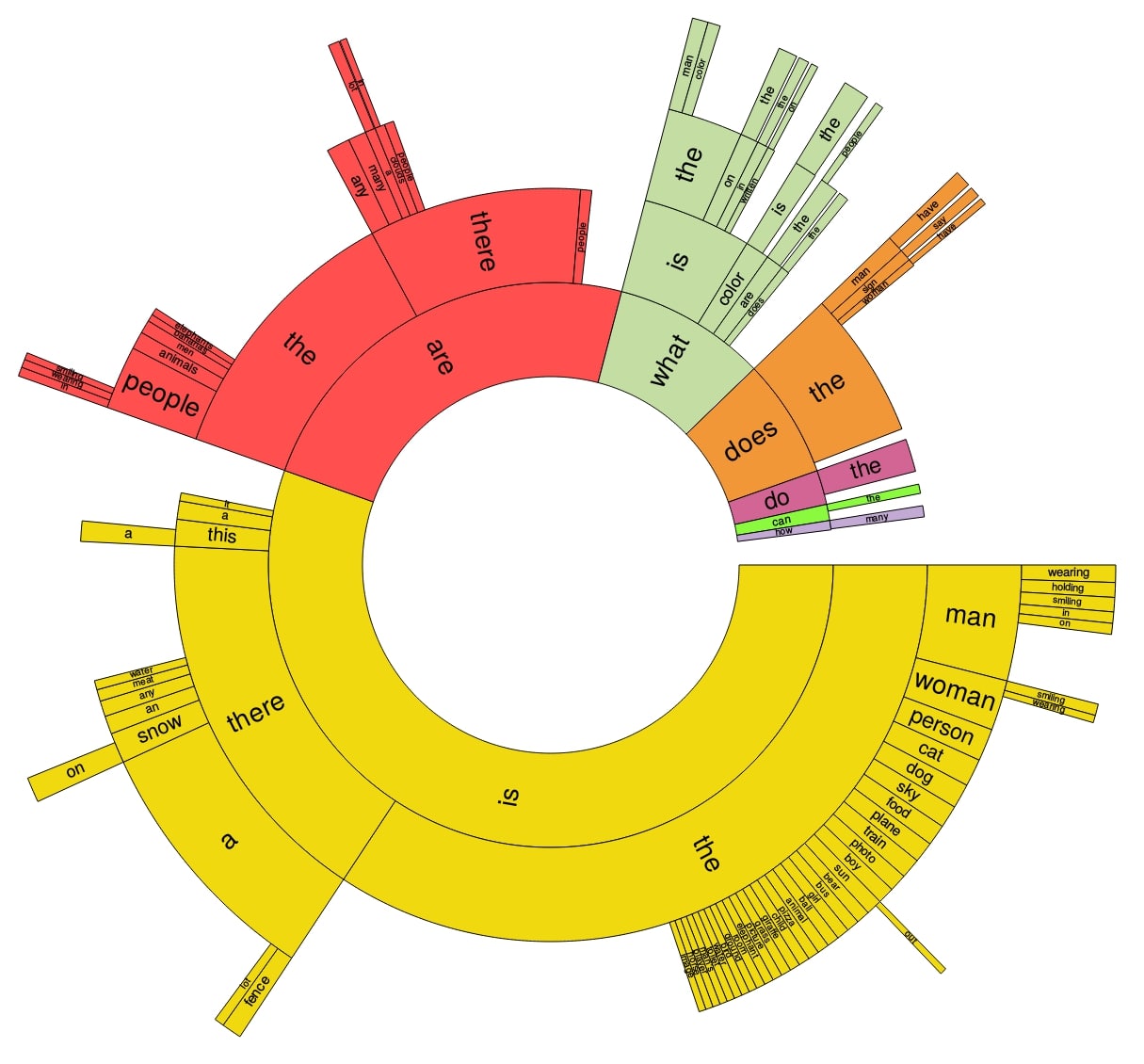}
   \label{fig:data_visual_a}
 \end{subfigure}
 \begin{subfigure}{.6\textwidth}
	 \centering
	 \vspace*{1cm}
	 \includegraphics[scale=0.12]{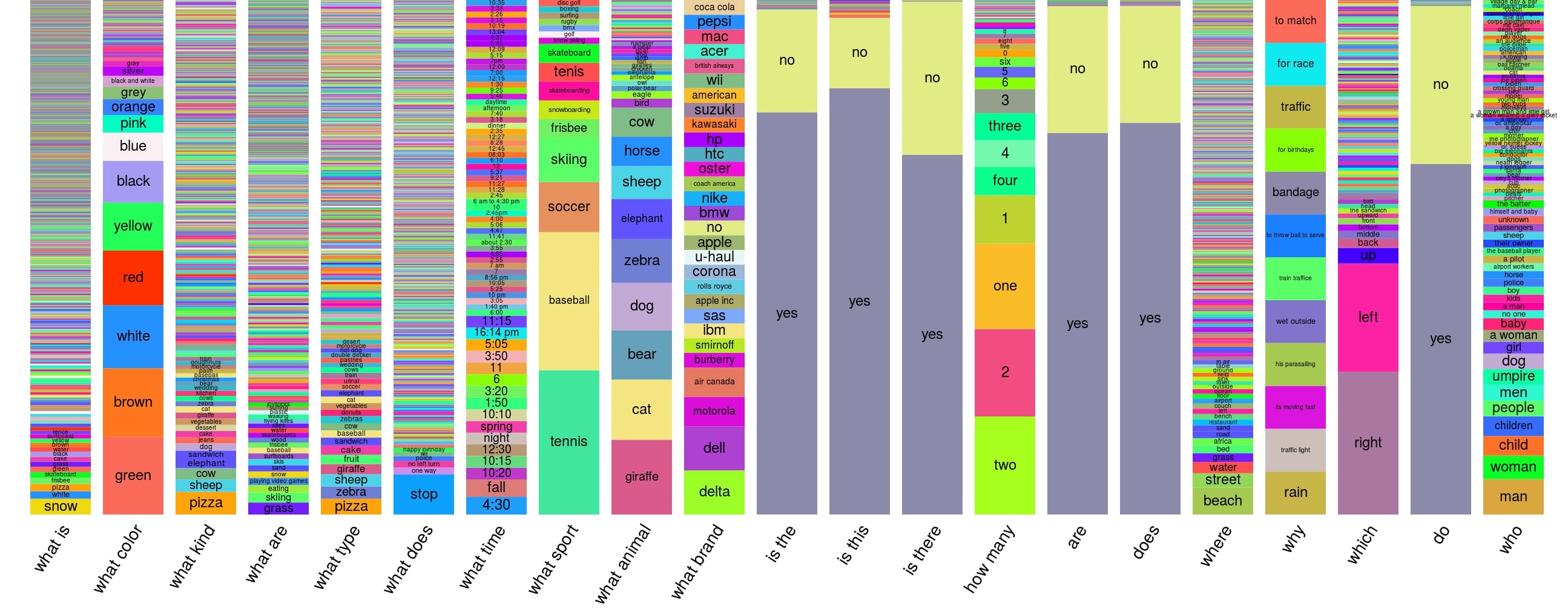}
	 \label{fig:data_visual_b}
 \end{subfigure}
\vspace{-10pt}
\caption{Left: Distribution of questions by their first four words. The arc length is proportional to the number of questions containing the word. White areas are words with contributions too small to show, Right: Distribution of answers per question type}
\vspace{\captionReduceBot}
\vspace{-5pt}
\label{fig:data_visual}
\end{figure*}

\begin{figure}[t]
\centering
\includegraphics[width=0.8\columnwidth]{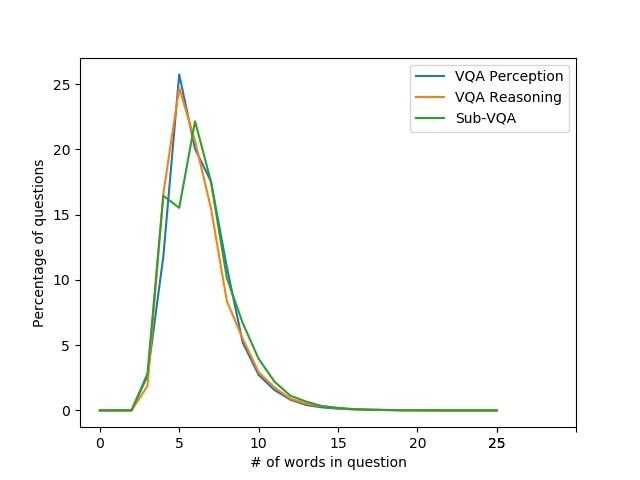}
\vspace{-5pt}
\caption{ Percentage of questions with different word lengths for the train and val sub-questions of our \data{} dataset.}
\vspace{\captionReduceBot}
\label{fig:len_subvqa}
\vspace{-12pt}
\end{figure}

The distribution of questions in our \data~dataset is shown in Figure~\ref{fig:data_visual}. 
It is interesting to note that comparing these plots with those for the VQA dataset \cite{antol2015vqa} show that the \data~dataset questions are more specific. 
For example, there are only 12 ``why'' questions in the dataset which tend to be reasoning questions. 
Also, for ``where'' questions, a very common answer in VQA was ``outside'' but answers are more specific in our \data~dataset (e.g., ``beach'', ``street''). 
Figure~\ref{fig:len_subvqa} shows the distribution of question lengths in the \perc~and \reas~splits of VQA and in our \data~dataset. We see that most questions range from 4 to 10 words. Lengths of questions in the \perc~and \reas~splits are quite similar, although questions in \data{} are slightly longer (the curve is slightly shifted to the right), possibly on account of the increase in specificity/detail of the questions.

One interesting question is whether the main question and the sub-questions deal with the same concepts. In order to explore this, we used noun chunks surrogates for concepts \footnote{Concepts are extracted with the Python spaCy library.}, and measured how often there was any overlap in concepts between the main question and the associated sub-question.
Noun-chunks are only a surrogate and may miss semantic overlap otherwise present (e.g. through verb-noun connections like ``fenced'' and ``a fence'' in Figure~\ref{fig:subq_data} (b), sub-questions). With this caveat, we observe that there is overlap only 24.18\% of the time, indicating that \reas~questions in our split often require knowledge about concepts not explicitly mentioned in the corresponding Perception questions. 
The lack of overlap indicates that models cannot solely rely on visual perception in answering Reasoning tasks, but incorporating background knowledge and common sense understanding is necessary. For example, 
in the question ``Is the airplane taking off or landing?'', the concepts present are `airplane' and `landing', while for the associated sub-question ``Are the wheels out?'', the concept is `wheels'. Though `wheels' do not occur in the main question, the concept is important, in that providing this grounding might help the model explicitly associate the connection between airplane wheels and take-offs / landings. 

\section{Fine grained evaluation of VQA \reas{}}
\label{consistency_eval}
\label{sec:eval}
\begin{figure*}[t!]
  \centering
  \includegraphics[scale=0.22]{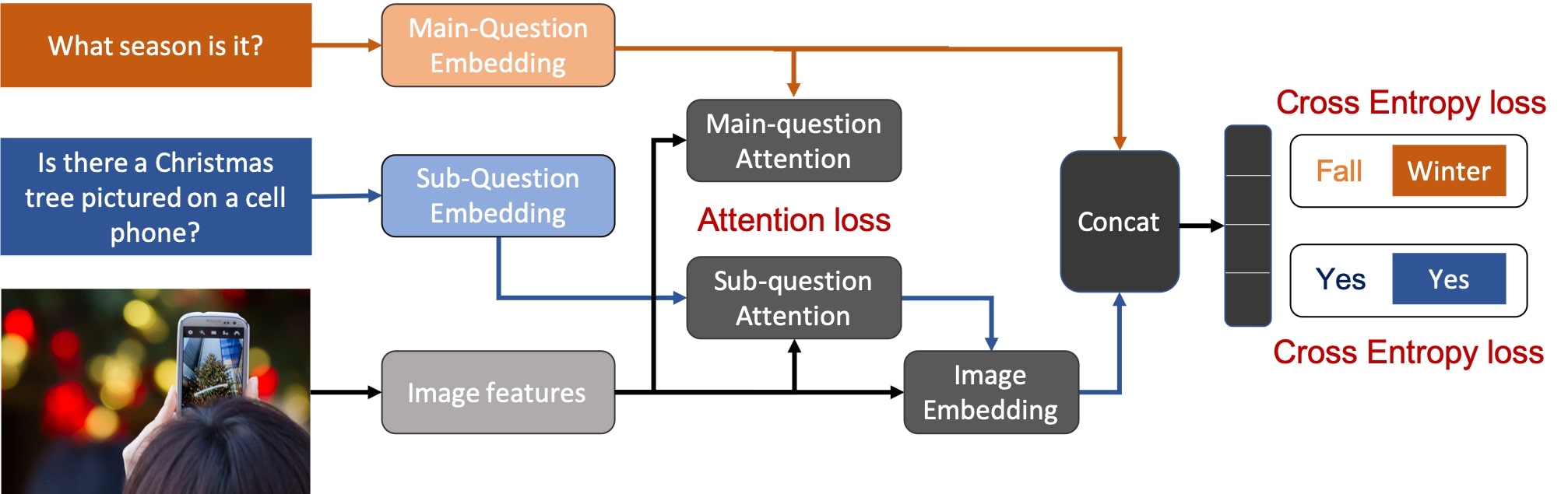}
  \vspace{-5pt}
   \caption{Sub-Question Importance-aware Network Tuning (SQuINT) approach: Given an image, a \reas~question like ``What season is it?'' and an associated \perc~sub-question like ``Is there a Christmas tree pictured on a cell phone?'', we pass them through the Pythia architecture \cite{jiang2018pythia}. The loss function customized for SQuINT is composed of three components: an attention loss that penalizes for the mismatch between attention for the main-question and the attention for the sub-question based on an image embedding conditioned on sub-question and image features, a cross entropy loss for answer of the main-question and a cross entropy loss for the answer of the sub-question. The loss function encourages the model to get the answers of both the main-question and sub-question right simultaneously, while also encouraging the model to use the right attention regions for the reasoning task.
   }
  \label{fig:squint_approach}
  \vspace{\captionReduceBot}
  \vspace{-5pt}
\end{figure*}

\vspace{-5pt}
\data{} enables a more detailed evaluation of the performance of current state-of-the-art models on \reas~questions by checking whether correctness on these questions is consistent with correctness on the associated \perc{} sub-questions. 
It is important to notice that a \perc{} failure (an incorrect answer to a sub-question) may be due to a problem in the vision part of the model or a grounding problem -- the model in Figure \ref{fig:squint_approach} may know that the banana is mostly yellow and use that information to answer the ripeness question, while, at the same time, fail to associate this knowledge with the word ``yellow'', or fail to understand what the sub-question is asking.
While grounding problems are not strictly visual perception failures, we still consider them \perc{} failures because the goal of VQA is to answer natural language questions about an image, and the sub-question being considered pertain to \perc{} knowledge as defined previously. With this caveat, there are four possible outcomes when evaluating \reas{} questions with associated \perc{} sub-questions, which we divide into four quadrants:

\noindent \colorbox{lightgreen}{\textbf{Q1: Both main \& sub-questions correct (M\cmark~S\cmark):}} While we cannot claim that the model predicts the main question correctly \emph{because} of the sub-questions (e.g. the bananas are ripe \emph{because} they are mostly yellow), the fact that it answers both correctly is consistent with good reasoning, and should give us more confidence in the original prediction.

\noindent\colorbox{lightblue}{\textbf{Q2: Main correct \& sub-question incorrect (M\cmark~S\xmark):}} The \perc{} failure indicates that there might be a reasoning failure. While it is possible that the model is composing other perception knowledge that was not captured by the identified sub-questions (e.g. the bananas are ripe because they have black spots on them), it is also possible (and more likely) that the model is using a spurious shortcut or was correct by random chance.

\noindent \colorbox{lightyellow}{\textbf{Q3: Main incorrect \& sub-question correct (M\xmark~S\cmark):}} The \perc{} failure here indicates a clear reasoning failure, as we validated that the sub-questions are sufficient to answer the main question. In this case, the model knows that the bananas are mostly yellow and still thinks they are not ripe enough, and thus it failed to make the ``yellow bananas are ripe'' connection.

\noindent \colorbox{lightpurple}{\textbf{Q4: Both main \& sub-question incorrect (M\xmark~S\xmark):}} While the model may not have the reasoning capabilities to answer questions in this quadrant, the \perc{} failure could explain the incorrect prediction. 

In sum, Q2 and Q4 are definitely \perc{} failures, Q2 likely contains \reas{} failures, Q3 contains \reas{} failures, and we cannot judge \reas{} in Q4.

As an example, we evaluate the Pythia model \cite{jiang2018pythia} (SOTA as of 2018)\footnote{source: \url{https://visualqa.org/roe_2018.html}} along these quadrants (Table \ref{tab:results}) for the \reas~split of VQA.
The overall accuracy of the model is 64.95\%, while accuracy on \reas{} questions is 69.61\%. We note that for 28.27\% of the cases,
the model is inconsistent, i.e., it answered the main question correctly, but got the sub question wrong.
Further, we observe that 9.02\% of the times the Pythia model gets \emph{all} the sub questions wrong when the main question is right -- \ie, it seems to be severely wrong on its perception and using other paths (shortcuts or biases) to get the Reasoning question right .

\vspace{-5pt}
\reducedSection{Improving learned models with \data{}}
\label{sec:approach}

In this section, we consider how \data{} can be used to improve models that were trained on VQA datasets. Our goal is to reduce the number of possible reasoning or perception failures (\colorbox{lightblue}{M\cmark~S\xmark}~and \colorbox{lightyellow}{M\xmark~S\cmark}) without diminishing the original accuracy of the model.
\vspace{-3pt}
\subsection{Finetuning} \label{finetuning}
\vspace{-3pt}
The simplest way to incorporate \data{} into a learned model is to fine-tune the model on it. 
We use the averaged binary cross entropy loss for the main question and the sub question as a loss function. 
Furthermore, to avoid catastrophic forgetting \cite{mccloskey1989catastrophic} of the  original VQA data during finetuning, we augment every batch with randomly sampled data from the original VQA dataset. 

\vspace{-5pt}
\subsection{Sub-Question Importance-aware Network Tuning (SQuINT)} \label{squint}
\vspace{-5pt}
The intuition behind Sub-Question Importance-aware Network Tuning (SQuINT) is that a model should attend to the same regions in the image when answering the \reas{} questions as it attends to when answering the associated \perc{} sub-questions, since they capture the visual components required to answer the main question.
SQuINT does this by learning how to attend to sub-question regions of interest and reasoning over them to answer the main question. 
We now describe how to construct a loss function that captures this intuition.

\noindent\textbf{Attention loss -}
As described in Section \ref{dataset}, the sub-questions in the dataset are simple perception questions asking about well-grounded objects/entities in the image. 
Current well-performing models based on attention are generally good at visually grounding regions in the image when asked about simple \perc~questions, given that they are trained on VQA datasets which contain large amounts of \perc~questions.
In order to make the model look at the associated sub-question regions while answering the main question, we apply a Mean Squared Error (MSE) loss over the the spatial and bounding box attention weights. 

\noindent\textbf{Cross Entropy loss -}
While the attention loss encourages the model to look at the right regions given a complex \reas{} question, we need a loss that helps the model learn to reason given the right regions. 
Hence we apply the regular Binary Cross Entropy loss on top of the answer predicted for the \reas~question given the sub-question attention. 
In addition we also use the Binary Cross Entropy loss between the predicted and GT answer for the sub-question. 

\noindent\textbf{Total SQuINT loss - }
We jointly train with the attention and cross entropy losses. %
Let $A_{reas}$ and $A_{sub}$ be the model attention for the main reasoning question and the associated sub-question, and $gt_{reas}$ and $gt_{sub}$ be the ground-truth answers for the main and sub-question respectively. Let $o_{reas}|A_{sub}$ be the predicted answer for the reasoning question given the attention for the sub-question. 
The SQuINT loss is formally defined as: 
\begin{align*}
\mathcal{L_{\textbf{SQuINT}}} &=   \textrm{MSE}(A_{reas}, A_{sub}) \\
&\quad + \lambda_{1}\textrm{BCE}(o_{reas}|A_{sub}, gt_{reas})\\ 
&\quad + \lambda_{2}\textrm{BCE}(o_{sub}, gt_{sub})
\vspace{-10pt}
\end{align*}

 \noindent The first term encourages the network to look at the same regions for reasoning and associated perception questions, while the second and third terms encourage the model to give the right answers to the questions given the attention regions. The loss is simple and can be applied as a modification to any model that uses attention.

\reducedSection{Experiments}

\begin{table*}[t] \footnotesize
\renewcommand*{\arraystretch}{1.3}
\setlength{\tabcolsep}{6pt}
\begin{center}
\resizebox{2\columnwidth}{!}{
\begin{tabular}{l l c  c c c c  c c c c c c}
\toprule 
& & & \multicolumn{7}{c}{Consistency Metric} & & \multicolumn{2}{c}{VQA Accuracy}\\
& Method & & \colorbox{lightgreen}{M\cmark~S\cmark} $\textcolor{OliveGreen}{\uparrow}$ & \colorbox{lightblue}{M\cmark~S\xmark} $\textcolor{red}{\downarrow}$ & \colorbox{lightyellow}{M\xmark~S\cmark} $\textcolor{red}{\downarrow}$ & \colorbox{lightpurple}{M\xmark~S\xmark} $\textcolor{red}{\downarrow}$ & Consistency\% $\textcolor{OliveGreen}{\uparrow}$ & Consistency\% (balanced) $\textcolor{OliveGreen}{\uparrow}$ & Attn Corr $\textcolor{OliveGreen}{\uparrow}$ & & Overall $\textcolor{OliveGreen}{\uparrow}$ & Reasoning (\colorbox{lightgreen}{M\cmark~S\cmark} + \colorbox{lightblue}{M\cmark~S\xmark}) $\textcolor{OliveGreen}{\uparrow}$ \\
\midrule
& Pythia && 50.05 & 19.73 & 17.40 & 12.83 & 71.73 & 75.67 & 0.71 & & \textbf{64.95} & 69.61 \\
& Pythia + \data{} data  && 53.21 & 16.62 & 19.40 & 10.77 & 76.20 & 74.58 & 0.71 & & 64.60 & 69.64 \\
\midrule
& Pythia + SQuINT && \textbf{53.90} & 16.24 & 19.34 & 10.52 & \textbf{76.84} & \textbf{75.76} & \textbf{0.75} & & 64.73 & \textbf{69.88} \\
\bottomrule
\end{tabular}}\\[5pt]
\vspace{-10pt}
\caption{Results on held out \data{} val set for (1) Consistency metrics along the four quadrants described in Section~\ref{consistency_eval} and Consistency and Attention Correlation metrics as described in Section~\ref{consistency_eval} (metrics), and (2) Overall and \reas~accuracy. 
The \reas~accuracy is obtained by only looking at the number of times the main question is correct (\colorbox{lightgreen}{M\cmark~S\cmark} + \colorbox{lightblue}{M\cmark~S\xmark}) ignoring repetitions of the main question due to multiple sub-questions. }
\label{tab:results}
\end{center}
\vspace{-10pt}
\end{table*}

\begin{figure*}[h!]
\vspace{-10pt}
 \centering
 \begin{subfigure}{.5\textwidth}
  \centering
  \includegraphics[scale=0.28]{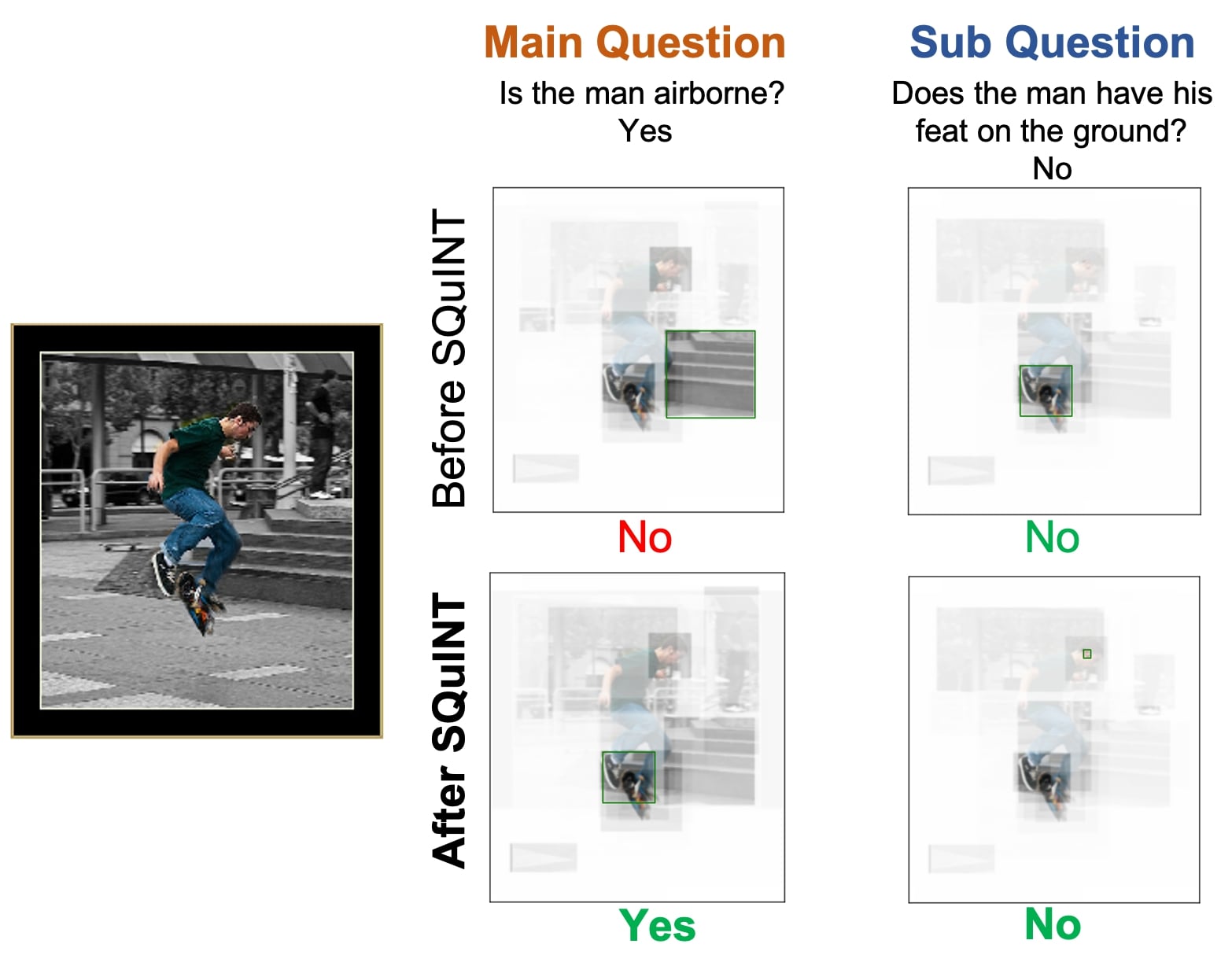}
  \vspace{-10pt}
  \caption{}
\end{subfigure}%
\begin{subfigure}{.5\textwidth}
  \centering
  \includegraphics[scale=0.14]{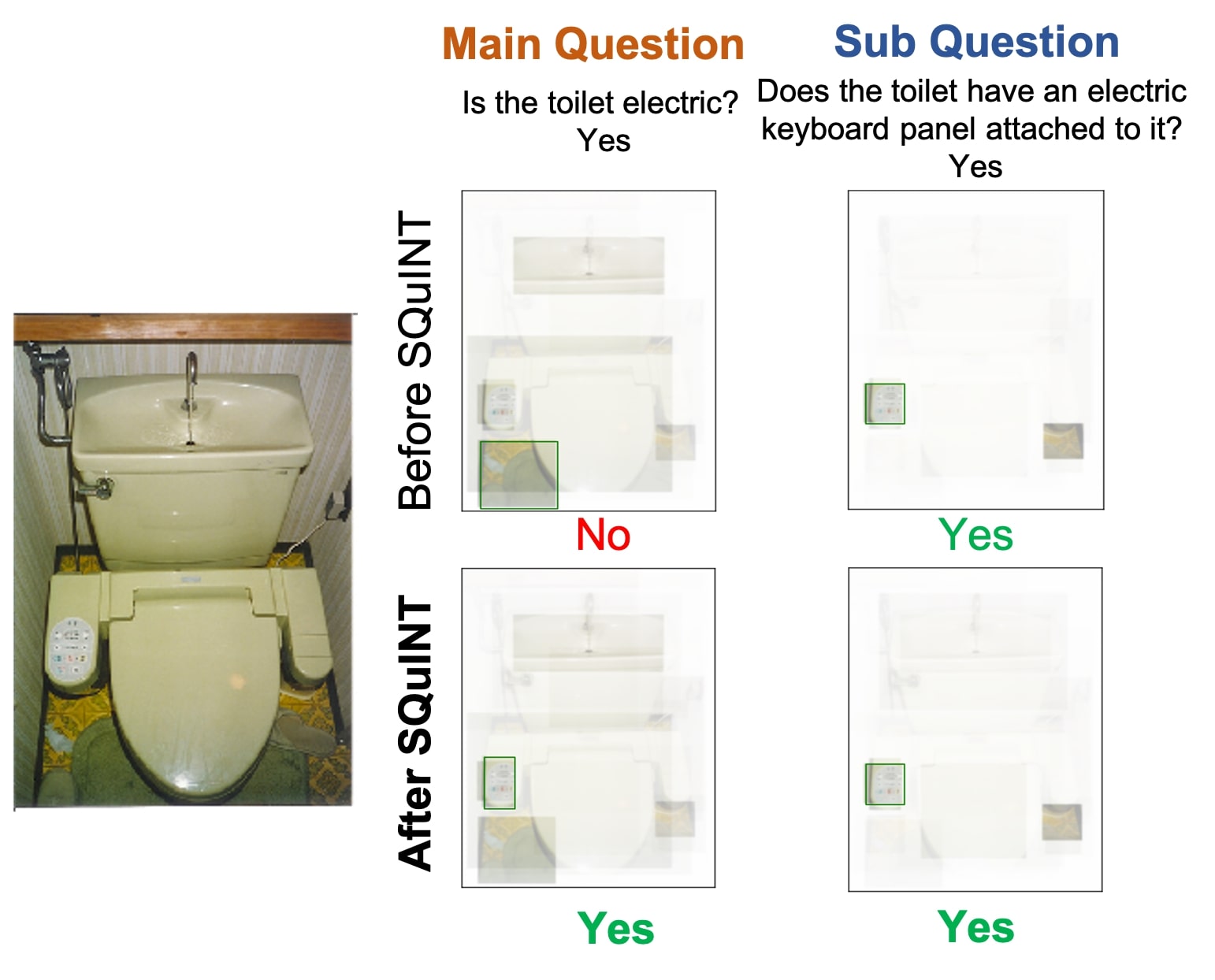}
  \vspace{-10pt}
  \caption{}
\end{subfigure}
\vspace{-10pt}
\caption{Qualitative examples showing the model attention before and after applying SQuINT. (a) shows an image along with the reasoning question, \myquote{Is the man airborne?}, for which the Pythia model looks at somewhat irrelevant regions and answers ``No" incorrectly. Note how the same model correctly looks at the feet to answer the easier sub-question, \myquote{Does the man have his feat on the ground?}. After applying SQuINT, which encourages the model to use the perception based sub question attention while answering the reasoning question, it now looks at the feet and correctly answers the main question.
\vspace{-15pt}
}
\label{fig:qual_squint}

\end{figure*}

In this section, we perform fine grained evaluation of VQA reasoning as detailed in Section \ref{sec:eval}, using the SOTA model \textbf{Pythia} \cite{jiang2018pythia} as a base model (although any model that uses visual attention would suffice).
We trained the base model on VQAv2, and evaluated the baseline and all variants on the \reas{} split and corresponding \data{} val sub-questions\footnote{Note that this is different from our CVPR camera-ready version where the experiments were conducted on the previous version of the \data{} dataset with a different base model that was trained on VQAv1.}. 
As detailed in Section \ref{sec:approach}, \textbf{Pythia + \data{} data} corresponds to finetuning the base model on train \data{} subquestions, while \textbf{Pythia + SQuINT} finetunes Pythia model such that it now attends to the same regions for main questions and associated sub-questions. 
In Table \ref{tab:results}, we report the reasoning breakdown detailed in Section \ref{sec:eval}. 
We also report a few additional metrics: \textbf{Consistency} refers to how often the model predicts the sub-question correctly given that it answered the main question correctly, while \textbf{Consistency (balanced)} reports the same metric on a balanced version of the sub-questions (to make sure models are not exploiting biases to gain consistency). 
\textbf{Attention Correlation} refers to the correlation between the attention embeddings of the main and sub-question. Finally, we report \textbf{Overall} accuracy (on the whole evaluation dataset), and accuracy on the Reasoning split (\textbf{Reasoning Accuracy}).
Note that our approach does not require sub-questions at test time. We use $\lambda_1=0.1$ and $\lambda_2=1$ with a learning rate of $0.01$ in our experiments.

The results in Table \ref{tab:results} indicate that fine-tuning on \data{} (using data augmentation or SQuINT), increases consistency without hurting accuracy or Reasoning accuracy. 
Correspondingly, our confidence that it actually learned the necessary concepts when it answered \reas{} questions correctly should increase.

The \textbf{Attention Correlation} numbers indicate that SQuINT really is helping the model use the appropriate visual grounding (same for main-question as sub-questions) at test time. 
This effect does not seem to happen with naive finetuning on \data{}. 
We present qualitative validation examples in Figure~\ref{fig:qual_squint}, where the base model attends to irrelevant regions when answering the main question (even though it answers correctly), while attending to relevant regions when asked the sub-question. 
The model finetuned on SQuINT, on the other hand, attends to regions that are actually informative in both main and sub-questions (notice that this is evaluation, and thus the model is not aware of the sub-question when answering the main question and vice versa). 
This is further indication that SQuINT is helping the model reason in ways that will generalize when it answers \reas{} questions correctly, rather than use shortcuts.

\vspace{-7pt}
\reducedSection{Discussion and Future Work}
The VQA task requires multiple capabilities in different modalities and at different levels of abstraction. 
We introduced a hard distinction between \perc{} and \reas{} which we acknowledge is a simplification of a continuous and complex reality, albeit a useful one. In particular, linking the perception components that are needed (in addition to other forms of reasoning) to answer reasoning questions opens up an array of possibilities for future work, in addition to improving evaluation of current work.
We proposed preliminary approaches that seem promising: fine-tuning on \data{} and SQuINT both improve the consistency of the SOTA model with no discernible loss in accuracy, and SQuINT results in qualitatively better attention maps. We expect future work to use \data{} even more explicitly in the modeling approach, similar to current work in explicitly composing visual knowledge to improve \emph{visual} reasoning \cite{hudson2018compositional}.
In addition, similar efforts to ours could be employed at different points in the abstraction scale, e.g. further dividing complex Perception questions into simpler components, or further dividing the \reas{} part into different forms of background knowledge, logic, etc. We consider such efforts crucial in the quest to evaluate and train models that truly generalize, and hope \data{} spurs more research in that direction.

\small{
\xhdr{Acknowledgements.}
We are grateful to Dhruv Batra for providing very useful feedback on this work. The Georgia Tech effort was supported in part by NSF, AFRL, DARPA, ONR YIPs, ARO PECASE, Amazon. The views and conclusions contained herein are those of the authors and should not be interpreted as necessarily representing the official policies or endorsements, either expressed or implied, of the U.S. Government, or any sponsor.}

\pagebreak
\begin{appendices}
\reducedSection{Introduction}

This supplementary material is organized as follows. 
We first provide a sample of the kind of regex-based rules that we used to arrive at reasoning questions. 
We then provide the interface we designed for training and evaluating Mechanical turk workers and the interface for collecting the main dataset. 
We then show randomly sampled responses from workers. 
\section{Perception-VQA vs Reasoning-VQA} \label{sup_rules}


In the first part of this section, we revisit our definition of \perc~and \reas~questions and later we describe our rules for constructing the \reas~split. 

\subsection{\perc~vs. \reas}

\noindent\textbf{\perc~:} As mentioned in section 3.1 of the main paper, we define \perc~questions as those which can be answered by detecting and recognizing the existence, physical properties and / or spatial relationships between entities, recognizing text / symbols, simple activities and / or counting, and that do not require more than one hop of reasoning or general commonsense knowledge beyond what is visually present in the image. 
Some examples are: ``Is that a cat? '' (existence), ``Is the ball shiny?'' (physical property), ``What is next to the table?'' (spatial relationship), ``What does the sign say?'' (text / symbol recognition), ``Are the people looking at the camera?'' (simple activity), etc. 

\noindent\textbf{\reas~:} We define \reas~questions as non-\perc~questions which require the synthesis of perception with prior knowledge and / or reasoning in order to be answered.
For instance, ``Is this room finished or being built?'', ``At what time of the day would this meal be served?'', ``Does this water look fresh enough to drink?'', ``Is this a home or a hotel?'', ``Are the giraffes in their natural habitat?'' are all \reas~ questions. 

\subsection{Rules}

As mentioned in section 3.1 of the main paper, our analysis of the perception questions in the VQA dataset revealed that most perception questions have distinct patterns that can be identified with high precision regex-based rules. 
In Table \ref{tab:rules} we provide a list of top-40 regex rules based on the percentage of data the rule eliminated. 

\begin{table*}[t] \footnotesize
\vspace{-23pt}
\renewcommand*{\arraystretch}{1.3}
\setlength{\tabcolsep}{6pt}
\begin{center}
\resizebox{2\columnwidth}{!}{
\begin{tabular}{c c c c  c c}
\toprule 
\multicolumn{4}{c}{Rules} & \multicolumn{2}{c}{Amount of Data} \\
Starts with & Contains & Not contains & Length & \# questions & \% data \\
\midrule
How many & - & - & - & 48656 & 10.96\\
- & color & - &  - & 47956 & 10.81\\
What is the & - & - &  - & 40988 & 9.24\\
What & on & - &  - & 29031 & 6.54\\
What & in & - &  - & 21876 & 4.93\\
Is there & - & - &  - & 16494 & 3.72 \\
- & wear & ['appropriate', 'acceptable', 'etiquitte']& - &  15530 & 3.50\\
- & wearing & - & - &  14940 & 3.37\\
Is this a & - & - & 4 &  14814 & 3.34\\
Where & - & - & - &  12409 & 2.80\\
- & old & - & - &  11197 & 2.52\\
What kind of & - & - & - & 11186 & 2.52\\
What are & - & - & - & 10524 & 2.37\\
- & on? & - & - & 9040 & 2.04\\
Are there & - & - & - & 8665 & 1.95\\
What type of & - & - & - & 7955 & 1.79\\
- & doing? & - & - &  7288 & 1.64\\
- & holding & - & - &  7137 & 1.61\\
- & low & - & - &  6596 & 1.49\\
- & round? & - & -  & 6242 & 1.41\\
Do & have & - & -  & 6213 & 1.40\\
Is the & on the & - & -  & 5375 & 1.21\\
Are these & - & ['homemade', 'healthy', 'domesticated', etc.] & 3 & 5320 & 1.20\\
Is the & in the & ['wild', 'mountain', 'desert', 'woods', etc.] & -  & 5108 & 1.15\\
Does & have & - & -  & 5078 & 1.14\\
- & number & - & -  & 4477 & 1.01\\
What is this & - & - & -  & 3970 & 0.89\\
Is & ed? & ['overexposed?', 'doctored?', 'ventilated?', etc.] & 3 & 3940 & 0.88\\
Is & ing? & ['horrifying?', 'relaxing?', 'competing?', etc.] & 3 & 3870 & 0.88\\
Is & on & - & 3 & 3622 & 0.82\\
Who & on & - & - & 3563 & 0.80\\
- & shown? & - & - & 3501 & 0.79\\
What sport & - & - & - & 3412 & 0.77\\
- & sun & - & - & 3260 & 0.73\\
- & see & - & - & 3238 & 0.73\\
- & visible & - & - & 3076 & 0.69\\
What & say? & - & - & 3238 & 0.69\\
What & playing? & - & - & 3076 & 0.69\\
Are the & in the &  ['US', 'wild', 'team', 'or', etc.] & - & 3010 & 0.68\\
What & playing? & - & - & 3076 & 0.69\\
Are & on the & - & - & 2932 & 0.66\\
\bottomrule
\end{tabular}}\\[5pt]
\vspace{-15pt}
\caption{Our rules for eliminating perception questions. Length refers to the $\#$ words in the question. We show top-40 rules based on data eliminated.}
\label{tab:rules}
\end{center}
\vspace{-30pt}
\end{table*}

\begin{figure*}[t!]

 \centering
 \begin{subfigure}{.5\textwidth}
  \centering
  \includegraphics[scale=0.12]{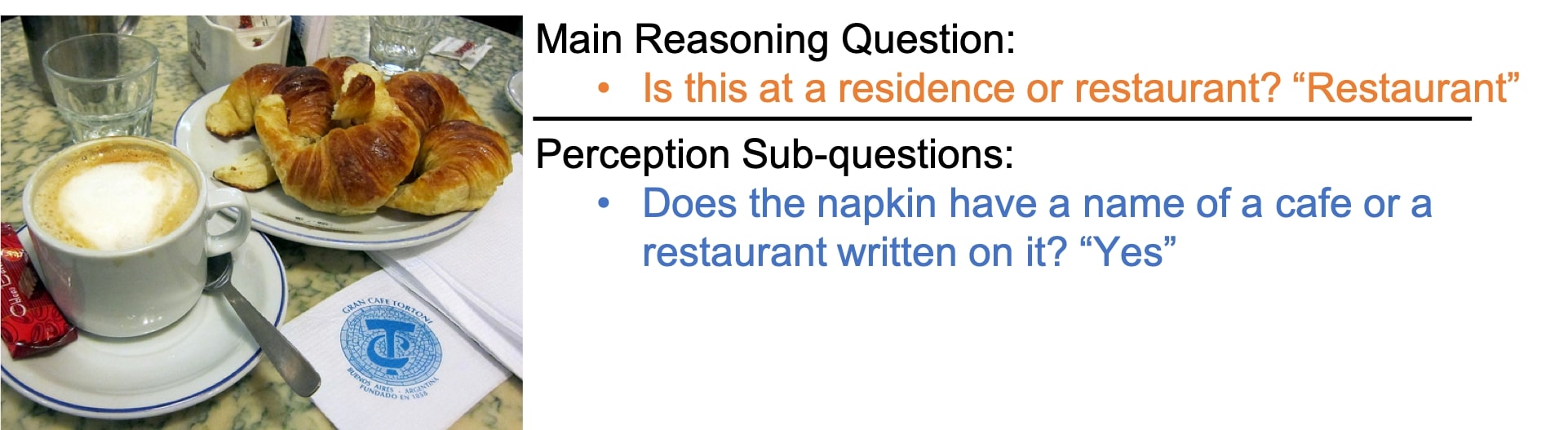}
  \caption{}
\end{subfigure}%
\begin{subfigure}{.5\textwidth}
  \centering
  \includegraphics[scale=0.12]{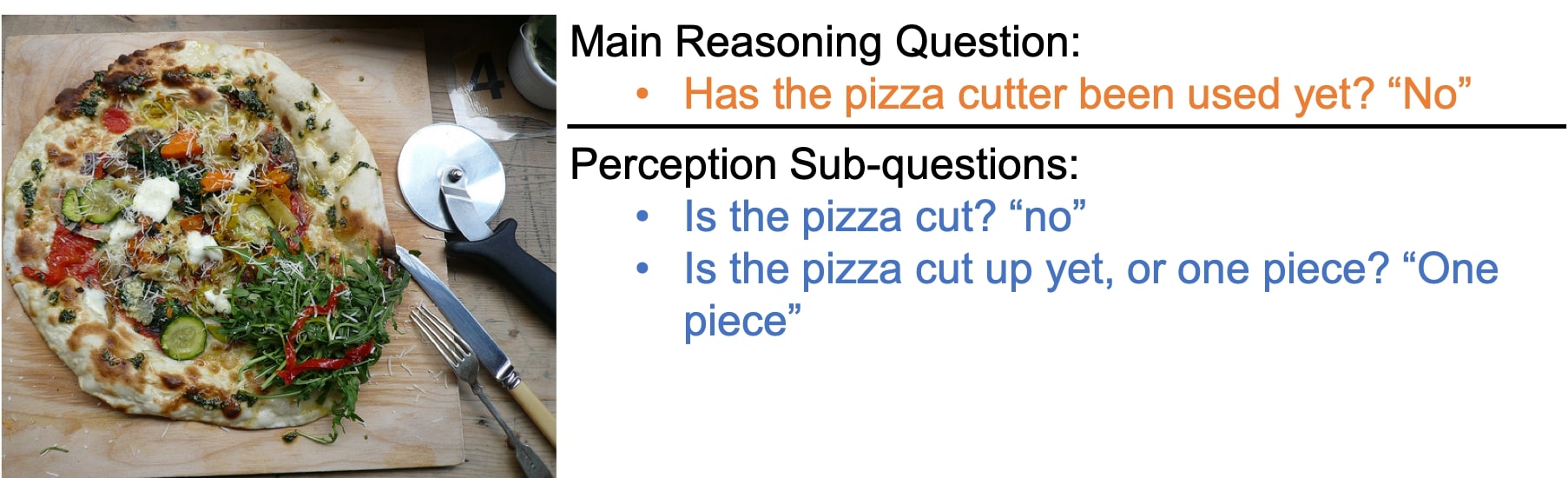}
  \caption{}
\end{subfigure}
\begin{subfigure}{.5\textwidth}
  \centering
  \includegraphics[scale=0.12]{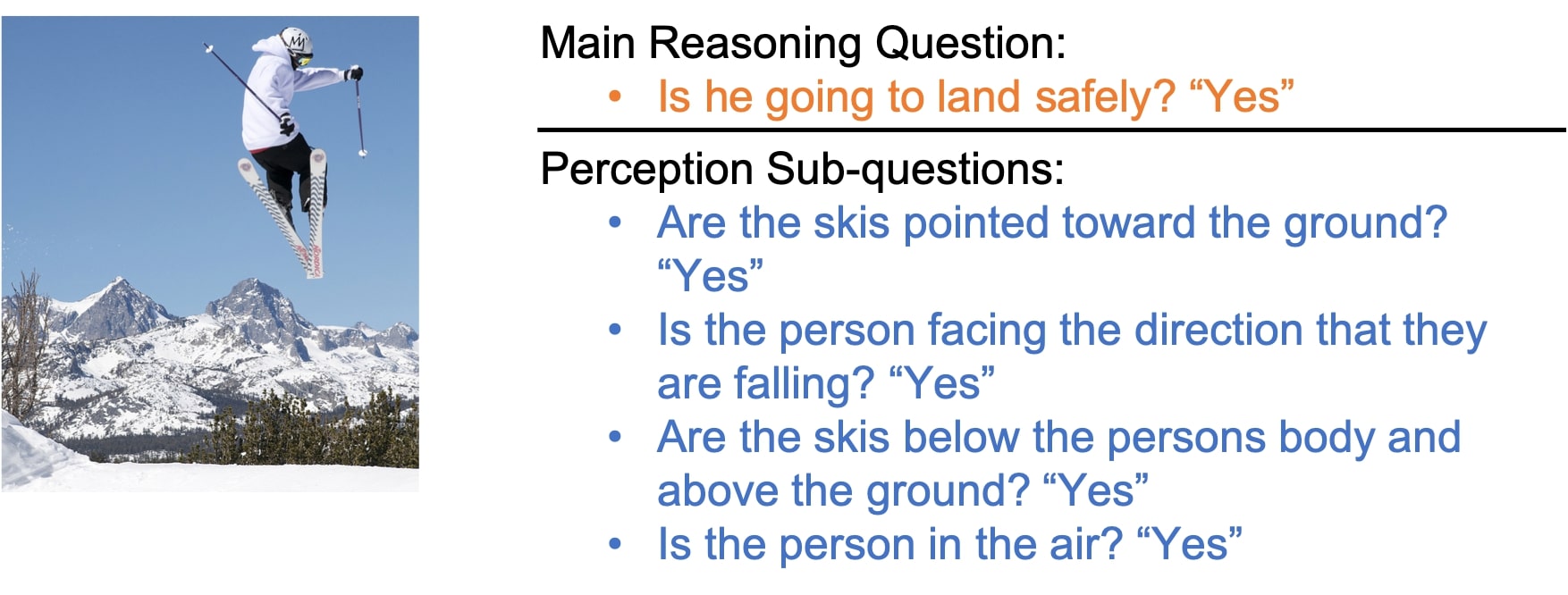}
  \caption{}
\end{subfigure}%
\begin{subfigure}{.5\textwidth}
  \centering
  \includegraphics[scale=0.12]{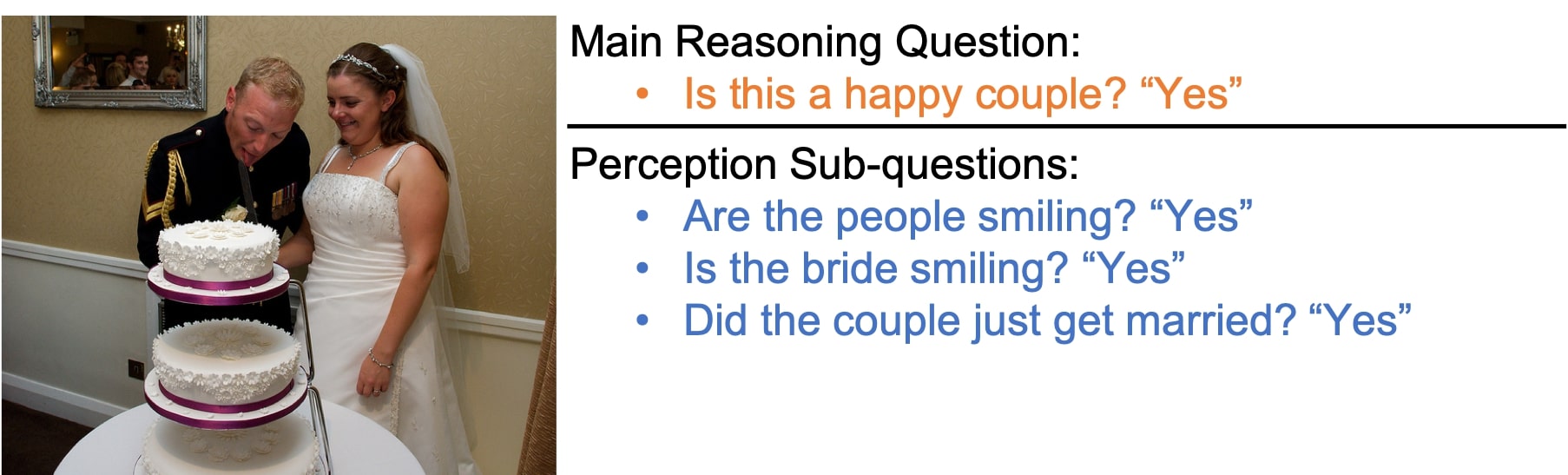}
  \caption{}
\end{subfigure}
\begin{subfigure}{.5\textwidth}
  \centering
  \includegraphics[scale=0.12]{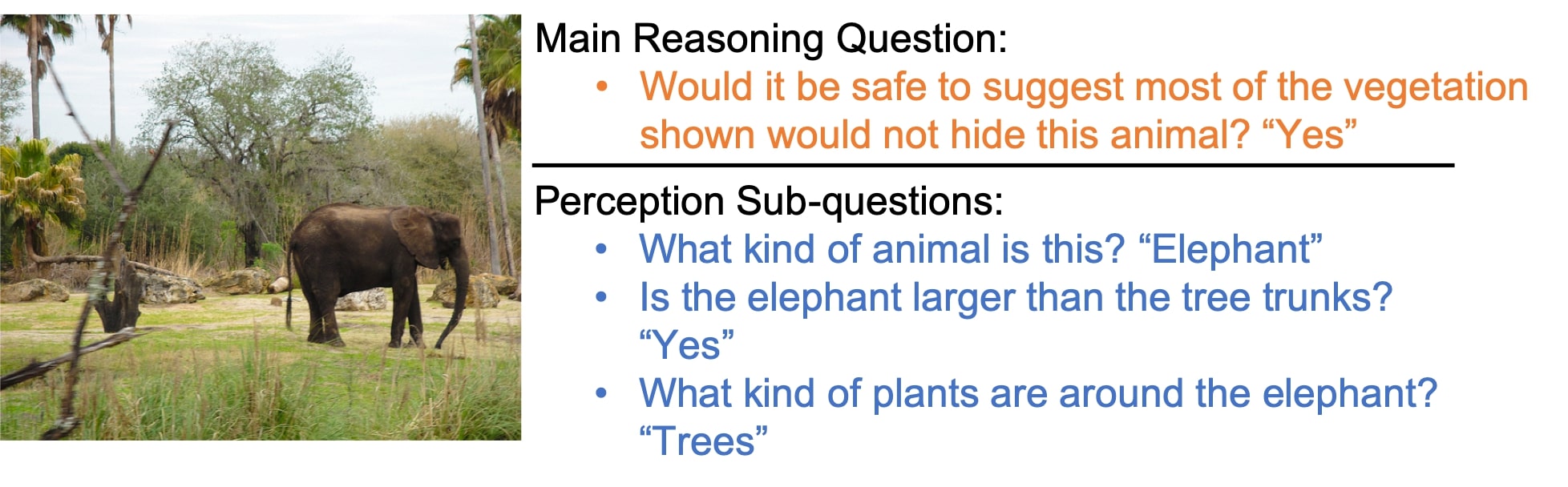}
  \caption{}
\end{subfigure}%
\begin{subfigure}{.5\textwidth}
  \centering
  \includegraphics[scale=0.12]{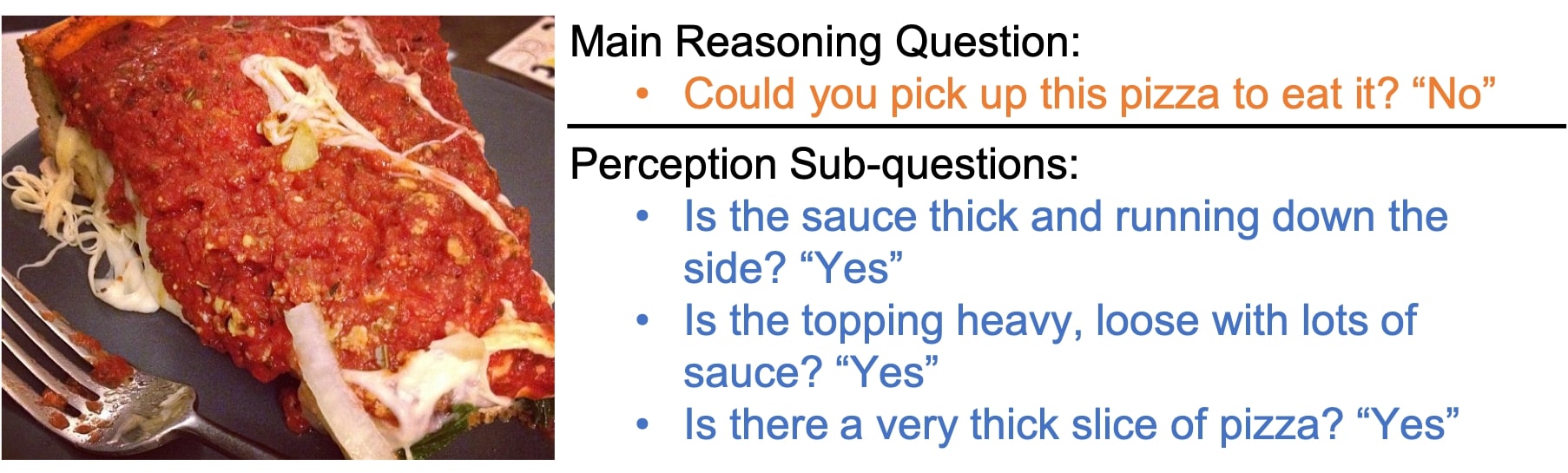}
  \caption{}
\end{subfigure}
\begin{subfigure}{.5\textwidth}
  \centering
  \includegraphics[scale=0.12]{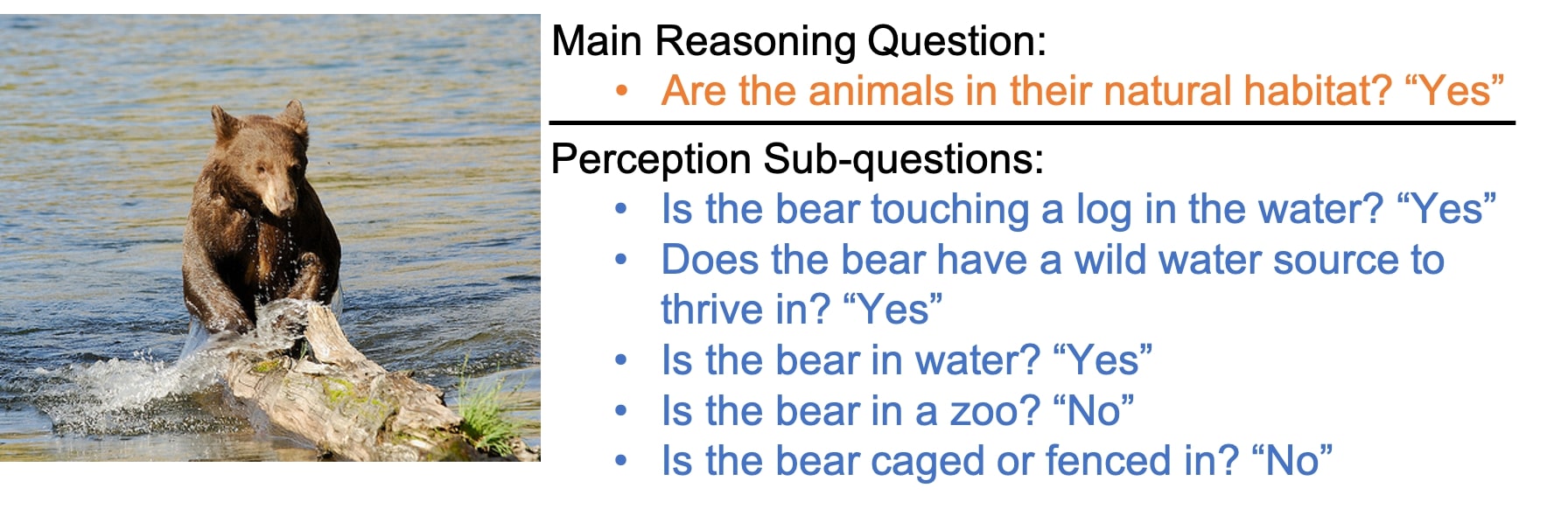}
  \caption{}
\end{subfigure}%
\begin{subfigure}{.5\textwidth}
  \centering
  \includegraphics[scale=0.12]{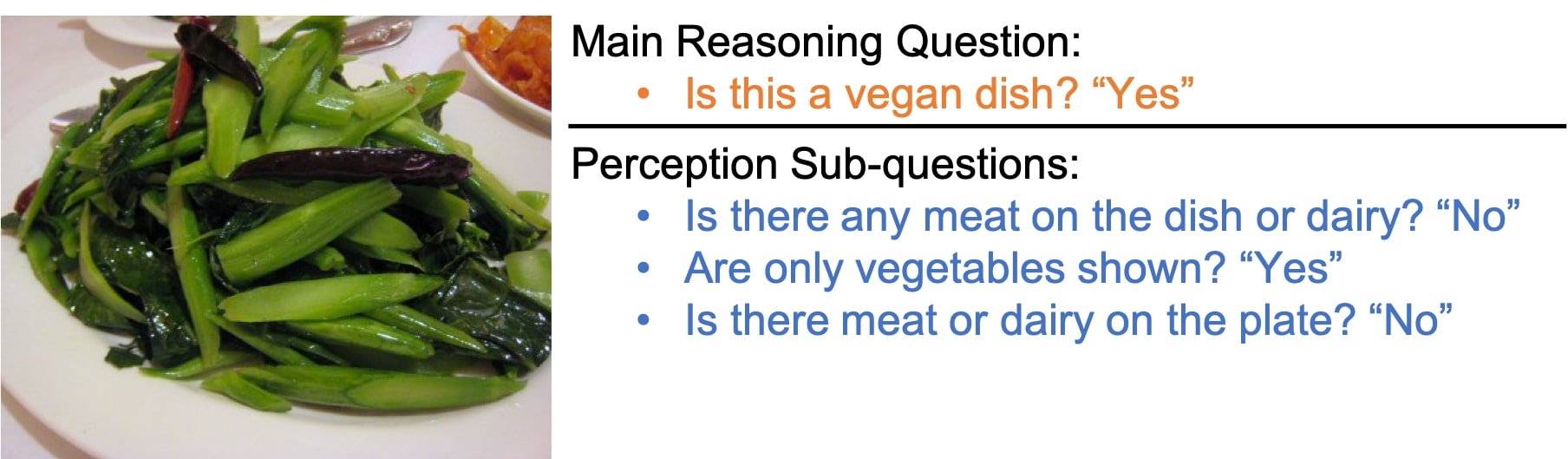}
  \caption{}
\end{subfigure}
\begin{subfigure}{.5\textwidth}
  \centering
  \includegraphics[scale=0.12]{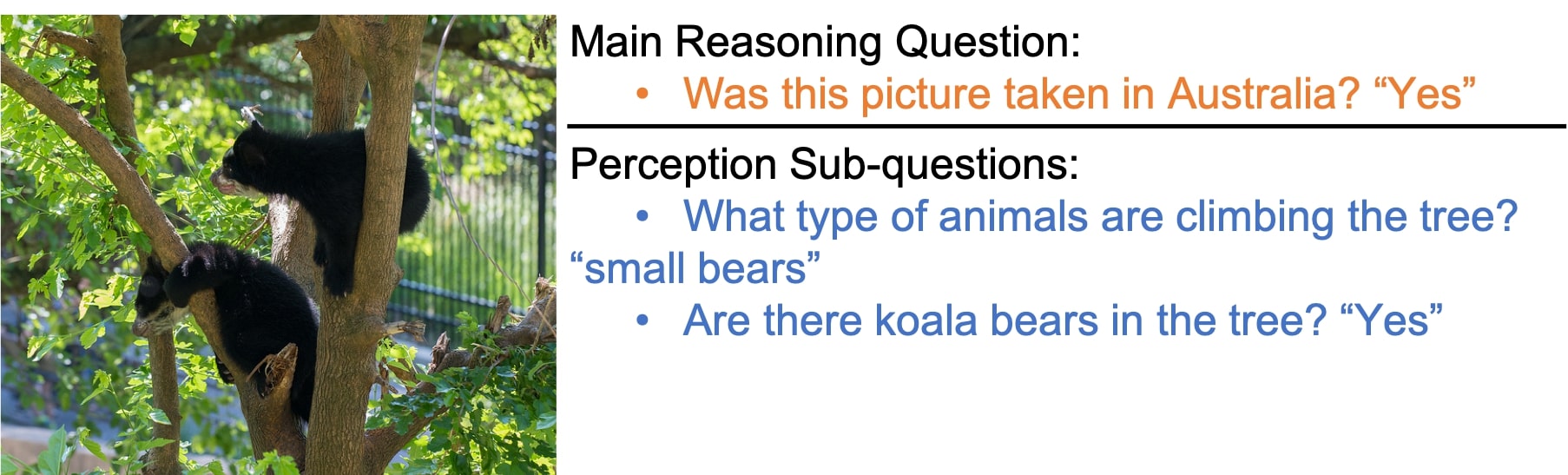}
  \caption{}
\end{subfigure}%
\begin{subfigure}{.5\textwidth}
  \centering
  \includegraphics[scale=0.12]{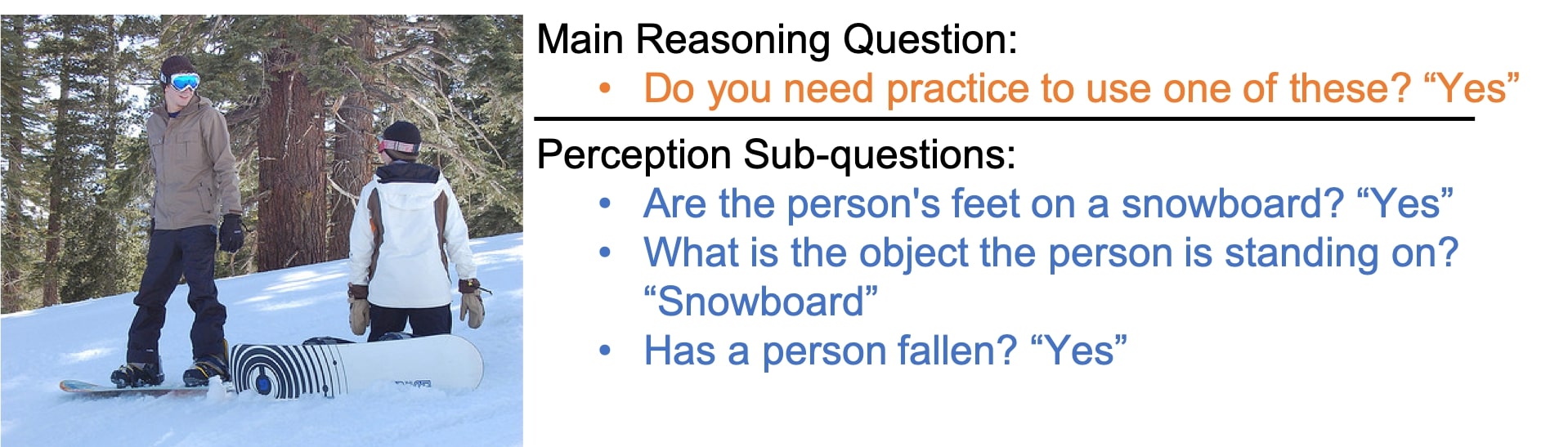}
  \caption{}
\end{subfigure}
\begin{subfigure}{.5\textwidth}
  \centering
  \includegraphics[scale=0.12]{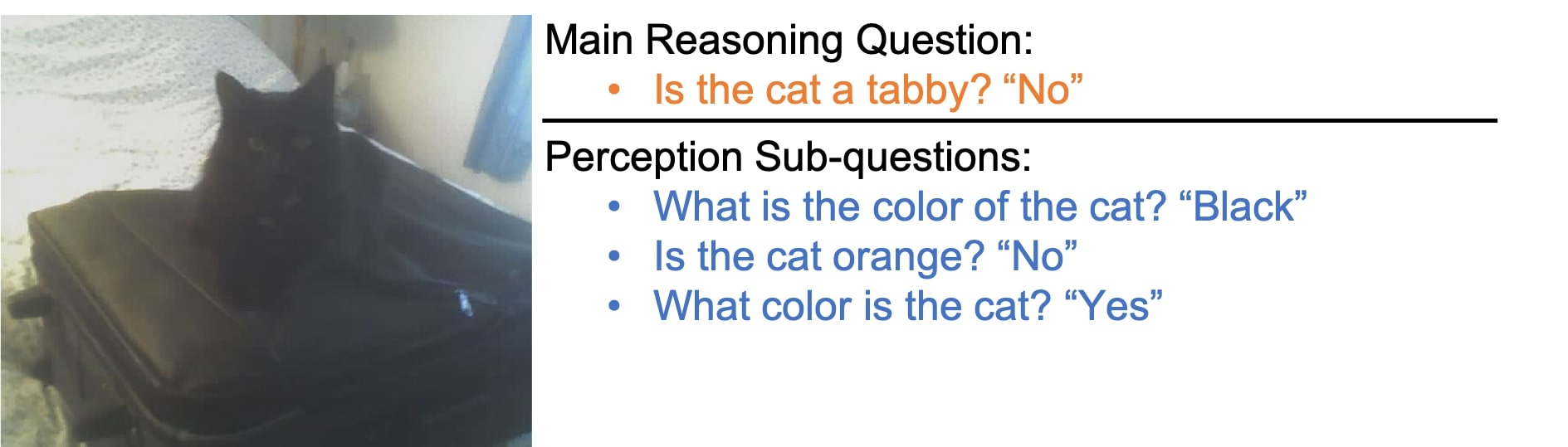}
  \caption{}
\end{subfigure}%
\begin{subfigure}{.5\textwidth}
  \centering
  \includegraphics[scale=0.12]{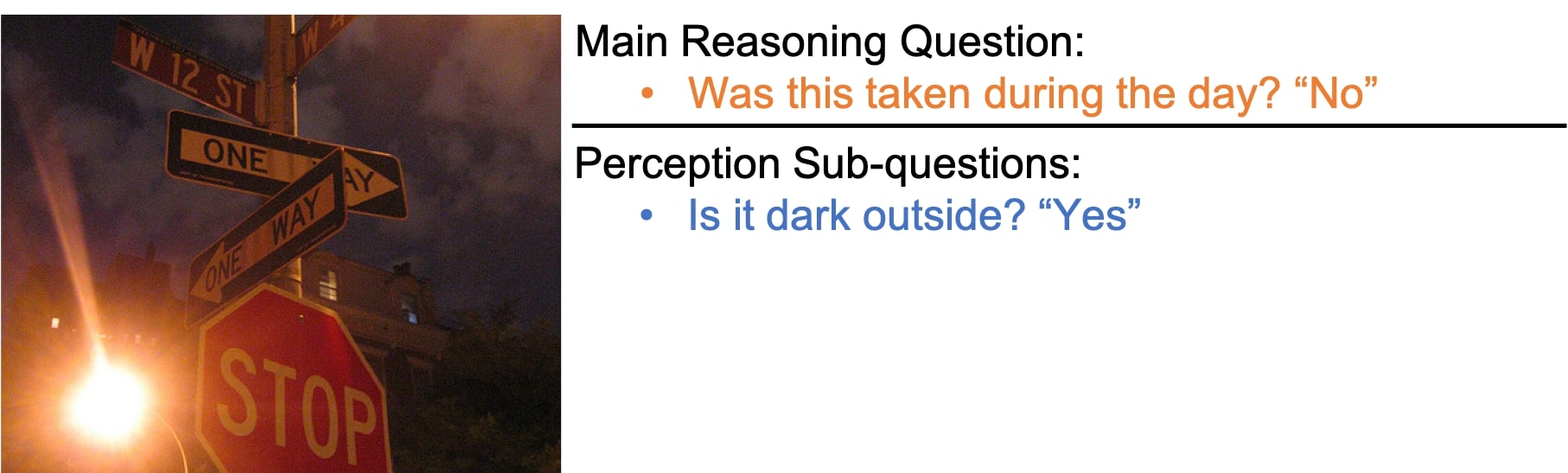}
  \caption{}
\end{subfigure}
\begin{subfigure}{.5\textwidth}
  \centering
  \includegraphics[scale=0.12]{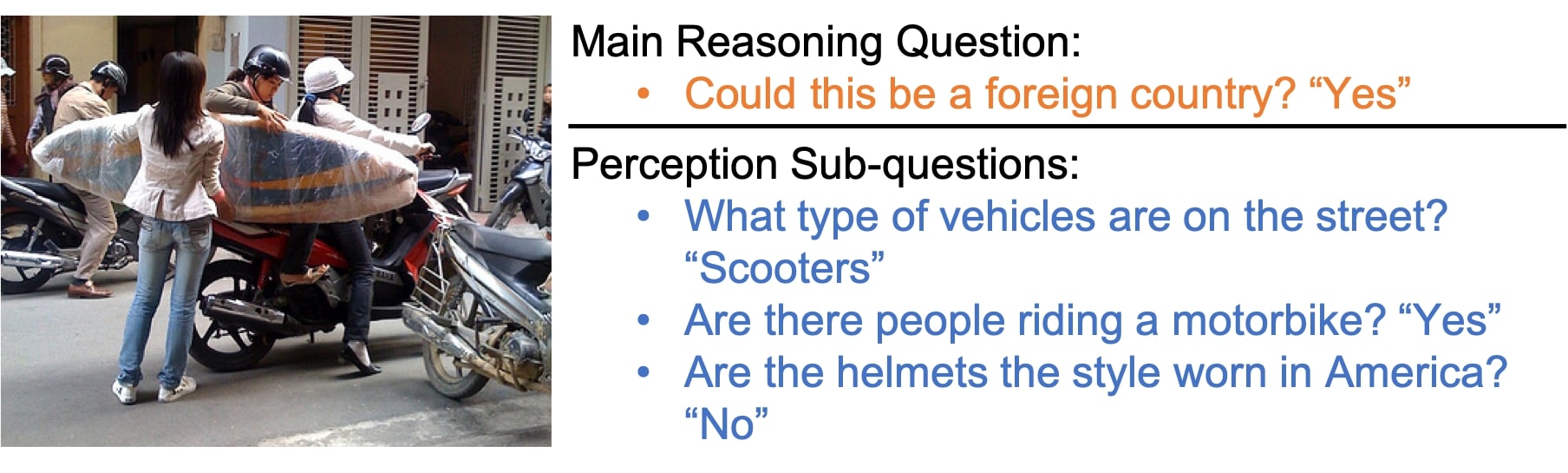}
  \caption{}
\end{subfigure}%
\begin{subfigure}{.5\textwidth}
  \centering
  \includegraphics[scale=0.12]{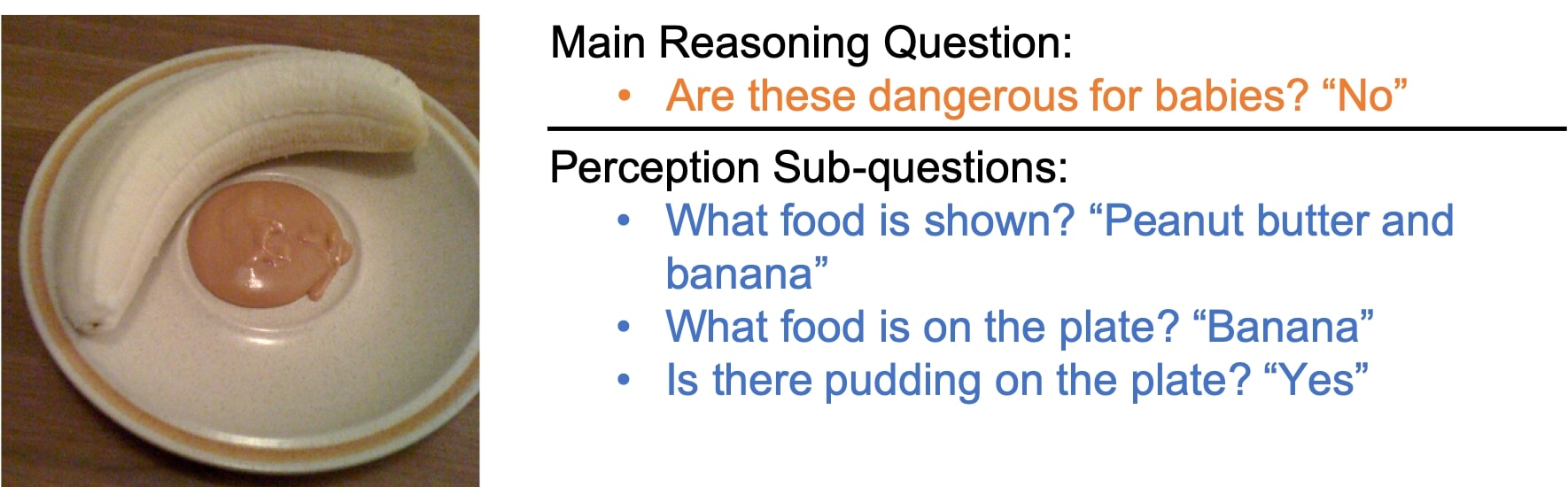}
  \caption{}
  \end{subfigure}
\caption{Randomly sampled qualitative examples of \perc~sub-questions in our \data{} dataset for main questions in the \reas~split of VQA. Main questions are written in \textcolor{orange}{orange} and sub questions are in \textcolor{blue}{blue}. A single worker may have provided more than one sub questions for the same (image, main question) pair.}
\vspace{-10pt}
\label{fig:subq_data}
\end{figure*}

By hand-crafting such rules (as seen in Table \ref{tab:rules}) and filtering out perception questions, we identify ~18\% of the VQA dataset as highly likely to be \reas. 

\subsection{Validating rules}

To check the accuracy of our rules, we designed a crowdsourcing task on Mechanical Turk that instructed workers to identify a given VQA question as \perc~or \reas, and to subsequently provide sub-questions for the \reas~questions. 

\xhdr{Validating Precision. }
As mentioned in Section 3.1, 89.25\% of the times, 2 out of 3 trained workers classified our resulting questions as reasoning questions demonstrating the high precision of the regex-based rules we created. 
\section{\data{}} \label{sup_dataset_collection}

In this section, we describe how we collect sub-questions and answers for questions in our \reas~split. 

Given the complexity of distinguishing between Perception / Reasoning and providing sub-questions for Reasoning questions, we first train and filter workers on Amazon Mechanical Turk (AMT) via qualification rounds before we rely on them to generate high-quality sub-questions.

\noindent\textbf{Worker Training - } \label{worker-training} We manually annotate $100$ questions from the VQA dataset as \perc{} and $100$ as \reas{} questions, to serve as examples.
We first teach workers the difference between \perc~and \reas~questions by defining them and showing several examples of each, along with explanations. 
Then, workers are shown (question, answer) pairs and are asked to identify if the given question is a \perc~question or a \reas~question \footnote{We also add an ``Invalid'' category to flag nonsensical questions or those which can be answered without looking at the image}. 
Finally, for \reas~ questions, we ask workers to add all \perc~questions and corresponding answers (in short) that would be necessary to answer the main question. 
In this qualification HIT, workers have to make 6 \perc{} and \reas{} judgments, and they qualify if they get 5 or more answers right.  
This interface can be found \href{https://filebox.ece.vt.edu/~ram21/SQuINT/qual1_interface.html}{here}. 

We launched further pilot experiments for the workers who passed the first qualification round, where we manually evaluated the quality of their sub-questions based on 2 criteria : (1) The sub-questions should be \perc~questions grounded in the image, and 2) The sub-questions should be sufficient to answer the main \reas~question. 
Among those 540 workers who passed the first qualification test, 144 were selected (via manual evaluation) as high-quality workers, which finally qualified for attempting our main task. 

\noindent\textbf{Main task - } 
In the main data collection, all VQA questions that got identified as \reas~by regex-rules (section \ref{sup_rules}) and a random subset of the questions identified as \perc~were further judged by workers (for validation purposes).  
We eliminated ambiguous questions by further filtering out questions where there is high worker disagreement about the answer. We require at least 8 out of 10 workers to agree with the majority answer for yes/no questions and 5 out of 10 for all other questions, which leaves us with a split that corresponds to $\sim$13\% of the VQA dataset. This interface can be found \href{https://filebox.ece.vt.edu/~ram21/SQuINT/main_interface.html}{here}. 

\subsection{\data{}} 

Each $<$question, image$>$ pair labeled as \reas~had sub questions generated by by 3 unique workers \footnote{A small number of workers displayed degraded performance after the qualification round, and were manually filtered}. On average we have 2.60 sub-questions per \reas~question. 

Randomly sampled qualitative examples from our collected dataset are shown in Fig. \ref{fig:subq_data}.

\end{appendices}



{\small
\bibliographystyle{ieee_fullname}
\bibliography{egbib}
}

\end{document}